\documentclass{article} 
\usepackage{iclr2023_conference,times}

\usepackage{hyperref}
\usepackage{url}
\usepackage{natbib}

\usepackage[utf8]{inputenc} 
\usepackage[T1]{fontenc}    
\usepackage{hyperref}       
\usepackage{url}            
\usepackage{booktabs}       
\usepackage{amsfonts}       
\usepackage{amsmath}
\usepackage{nicefrac}       
\usepackage{microtype}      

\usepackage{comment}
\usepackage{graphicx} 
\usepackage{subfigure} 
\usepackage{pdflscape}
\usepackage{amssymb}
\usepackage{algorithm}
\usepackage{algorithmic}
\usepackage{bm}
\usepackage{soul}
\usepackage{listings}
\usepackage{bbm}
\usepackage{multirow}
\usepackage{multicol}

\include{formatting}

\newcommand\Tstrut{\rule{0pt}{2.2ex}}         

\newcommand{\cut}[1]{}

\newcommand{\ourModel}{AGFA}
\newcommand{\ourModelLong}{Adversarial Generation of Fourier Amplitude}

\title{Domain Generalisation via Domain Adaptation: An Adversarial Fourier Amplitude Approach}

\author{Minyoung Kim$^1$, \ Da Li$^1$ \ \& \ Timothy M.~Hospedales$^{1,2}$ \\
$^1$Samsung AI Center Cambridge, UK \ \ \ \ \ \ \ \ \ \ \ \ \ \ \ \ \ \ \ \ \ \ \ \ \ \ \ \ \ \  
$^2$University of Edinburgh, UK \\
\texttt{\{mikim21,dali.academic\}@gmail.com \ \ \ \ t.hospedales@ed.ac.uk}
}

\iclrfinalcopy 
\begin{document}

\maketitle

\begin{abstract}
We tackle the domain generalisation (DG) problem by posing it as a domain adaptation (DA) task where we adversarially synthesise the worst-case `target' domain and adapt a model to that worst-case domain, thereby improving the model’s robustness. To synthesise data that is challenging yet semantics-preserving, we generate Fourier amplitude images and combine them with source domain phase images, exploiting the widely believed conjecture from signal processing that amplitude spectra mainly determines image style, while phase data mainly captures image semantics. To synthesise a worst-case domain for adaptation, we train the classifier and the amplitude generator adversarially. Specifically, we exploit the maximum classifier discrepancy (MCD) principle from DA that relates the target domain performance to the discrepancy of classifiers in the model hypothesis space. By Bayesian hypothesis modeling, we express the model hypothesis space effectively as a posterior distribution over classifiers given the source domains, making adversarial MCD minimisation feasible. On the DomainBed benchmark including the large-scale DomainNet dataset, the proposed approach yields significantly improved domain generalisation performance over the state-of-the-art.
\end{abstract}

\section{Introduction}
Contemporary machine learning models perform well when training and testing data are identically distributed. However, in practice it is often impossible to obtain an unbiased sample of real-world data for training, and therefore distribution-shift inevitably exists between training and deployment. Performance can degrade dramatically under such domain shift \citep{koh2021wilds}, and this is often the cause of poor performance of real-world deployments \citep{geirhos2020shortcut}. This important issue has motivated a large amount of research into the topic of domain generalisation (DG) \citep{zhou2021domainGenSurvey}, which addresses training models with increased robustness to distribution shift. These DG approaches span a diverse set of strategies including architectural innovations \citep{dmg}, novel regularisation \citep{metareg}, alignment \citep{coral} and learning  \citep{li2019episodic} objectives, and data augmentation~\citep{mixstyle} to make available training data more representative of potential testing data. However, the problem remains essentially unsolved, especially as measured by recent carefully designed benchmarks \citep{domainbed}.

\cut{Our approach is related to to existing lines of work on data augmentation solutions to DG \citep{mixstyle,shankar2018crossgrad}, which synthesise novel data for model training; and also builds on domain alignment strategies from the adjacent problem of domain adaptation \citep{coral,saito2018}. Augmentation-based approaches must specify solutions to two challenging issues: (1) How to synthesise data that is challenging enough to usefully augment source data, yet retains the task semantics so? and (2) How to use the synthesised data for learning? For example CrossGrad \citep{shankar2018crossgrad} synthesises data with a domain adversarial objective (data should be unlike any known domains), preserves semantics by requiring synthetic data to be no more than an $\epsilon$-perturbation different from existing training data, and simply adds the synthetic data to the training set. Our approach provides principled and novel answers to these questions. }

Our approach is related to existing lines of work on data-augmentation solutions to DG  \citep{mixstyle,shankar2018crossgrad}, which synthesise more data for model training; and alignment-based approaches to Domain Adaptation \citep{coral,saito2018} that adapt a source model to an unlabeled target set -- but cannot address the DG problem where the target set is unavailable. We improve on both by providing a unified framework for stronger data synthesis and domain alignment. 

\begin{figure}[t!]
\vspace{-3.0em}
\begin{center}
%
\centering
\includegraphics[trim = 4mm 4mm 4mm 4mm, clip, scale=0.325]{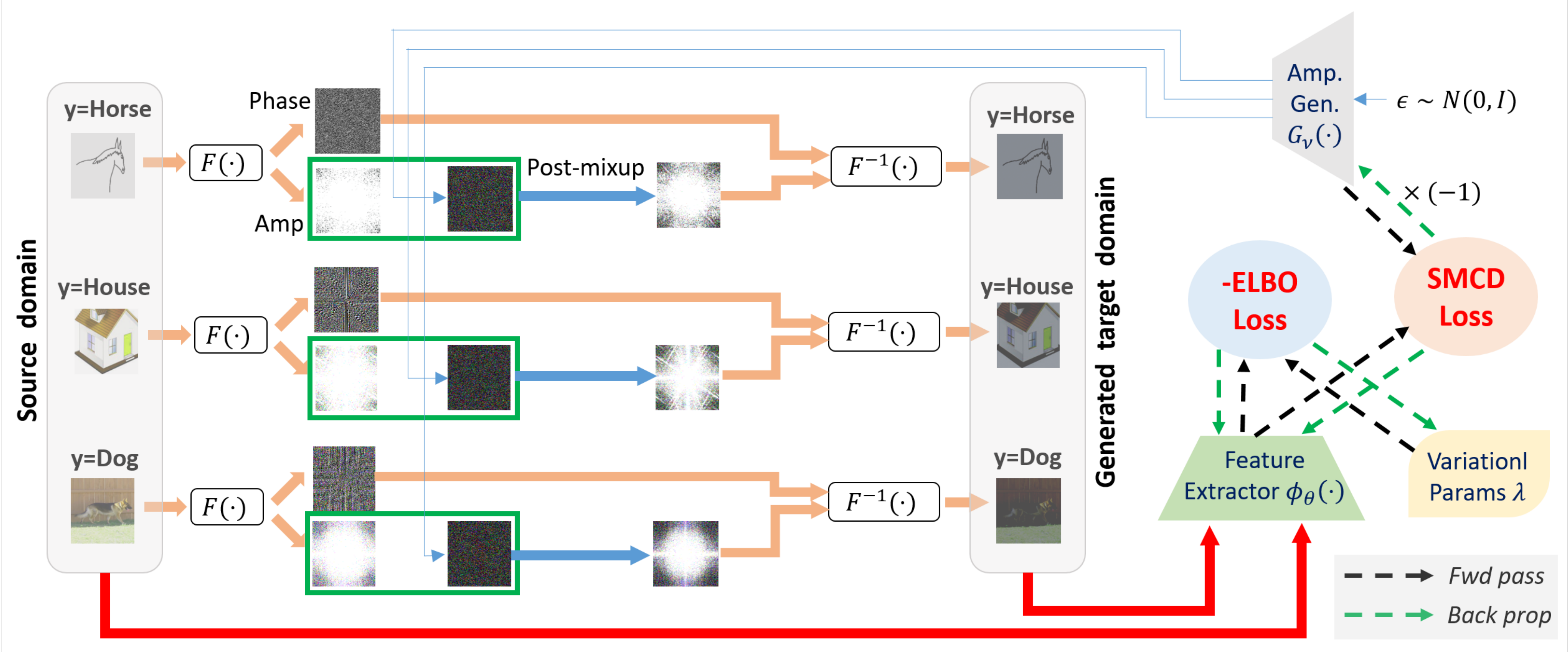}
\end{center}
\vspace{-1.3em}
\caption{Overall training flow of the proposed approach (AGFA). We generate target-domain data by synthesizing Fourier amplitude images 
trained adversarially. 
See main text in Sec.~\ref{sec:main} for details.
}
\vspace{-1.0em}
\label{fig:main}
\end{figure}

Our framework combines two key innovations: A Bayesian approach to maximum classifier discrepancy, and a Fourier analysis approach to data augmentation. We start from the perspective of maximum classifier discrepancy (MCD) from domain adaptation \citep{ben-david-2007,ben-david-2010,saito2018}. This bounds the target-domain error as a function of discrepancy between multiple source-domain classifiers. It is not obvious how to apply MCD to the DG problem where we have no access to target-domain data. A key insight is that MCD provides a principled objective that we can maximise in order to \emph{synthesise} a worst-case target domain, and also minimise in order to train a model that is adapted to that worst-case domain. Specifically, we take a Bayesian approach that learns a distribution over source-domain classifiers, with which we can compute MCD. This simplifies the model by eliminating the need for adversarial classifier training in previous applications of MCD \citep{saito2018}, which leaves us free to adversarially train the worst-case target domain. To enable challenging worst-case augmentations to be generated without the risk of altering image semantics, our augmentation strategy operates in the Fourier amplitude domain. It synthesises amplitude images, which can be combined with phase images from source-domain data to produce images that are substantially different in style (amplitude), while retaining the original semantics (phase). Our overall strategy termed \ourModelLong{} (\ourModel) is illustrated in  Fig.~\ref{fig:main}. 

In summary, we make the following main contributions: (1) We provide a novel and principled perspective on DG by drawing upon the MCD principle from DA. (2) We provide \ourModel{}, an effective algorithm for DG based on variational Bayesian learning of the classifier and Fourier-based synthesis of the worst-case domain for robust learning. (3) Our empirical results show clear improvement on previous state-of-the-arts on the rigorous DomainBed benchmark.

\section{Problem Setup and Background}\label{sec:setup}
\vspace{-0.5em}
We follow the standard setup for the Domain Generalisation (DG) problem. As training data, we are given labeled data $S = \{(x,y) \vert (x,y)\sim\mathcal{D}_i, i=1,\dots, N\}$ where $x\in\mathcal{X}$ and $y\in\mathcal{Y}=\{1,\dots,C\}$. Although the source domain $S$ consists of different domains $\{\mathcal{D}_i\}_{i=1}^N$ with domain labels available, we simply take their union without using the originating domain labels. This is because in practice the number of domains ($N$) is typically small, and it is rarely possible to estimate a meaningful population distribution for empirical $S$ from a few different domains. 
What distinguishes DG from the closely-related (unsupervised) Domain Adaptation (DA), is that the target domain ($T$) on which model's prediction performance is measured is {\em unknown} for DG, whereas in DA the input data $x$ from the target domain are revealed (without class labels $y$). 
Below we briefly summarise the MCD principle and Ben-David's theorem, one of the key theorems in DA, as we exploit them to tackle DG. 

\noindent\textbf{Ben-David's theorem and MCD principle in DA.}\quad In  unsupervised DA, Ben-David's theorem~\citep{ben-david-2010,ben-david-2007} provides an upper bound for the target-domain generalisation error of a model (hypothesis). We focus on the tighter bound version, which states that for any classifier $h$ in the hypothesis space $\mathcal{H}=\{h| h:\mathcal{X}\rightarrow\mathcal{Y}\}$, the following holds (without the sampling error term):
\begin{align}
e_T(h) \ \leq \  
  e_S(h) + \sup_{h,h' \in \mathcal{H}} 
    \big\vert d_S(h,h') - d_T(h,h') \big\vert  + e^*(\mathcal{H};S,T),
\label{eq:thm_2010}
\end{align}
where $e_S(h) := \mathbb{E}_{(x,y)\sim S} [ \mathbb{I}(h(x) \neq y) ]$ is the error rate of $h(\cdot)$ on the source domain $S$, $d_S(h,h') := \mathbb{E}_{x\sim S} [ \mathbb{I}(h(x) \neq h'(x)) ]$ denotes the discrepancy between two classifiers $h$ and $h'$ on 
$S$ (similarly for $e_T(h)$ and $d_T(h,h')$), 
and $e^*(\mathcal{H};S,T):= \min_{h\in\mathcal{H}} e_S(h) + e_T(h)$. 
Thus we can provably reduce the target domain generalisation error by simultaneously minimizing the three terms in the upper bound\footnote{Some recent work such as~\citep{investigate_dg}, however, empirically studied potential risk of looseness of the bound in certain scenarios.}, namely source-domain error $e_S(h)$, classifier discrepancy, and minimal source-target error. 

Previous approaches~\citep{saito2018,gpda} aim to minimise the upper bound, and one reasonable strategy is to {\em constrain} the hypothesis space $\mathcal{H}$ in such a way that it contains only those $h$'s with small $e_S(h)$. Within this source-confined hypothesis space (denoted by $\mathcal{H}_{|S}$), 
the terms $e_S(h)$ and $d_S(h,h')$ in the bound are expected to be 
close to $0$ for all $h,h'\in\mathcal{H}_{|S}$, and the bound of (\ref{eq:thm_2010}) effectively reduces to what is called the {\em Maximum Classifier Discrepancy} (MCD) loss,
\begin{equation}
\textrm{MCD}(\mathcal{H}_{|S};T) := 
\sup_{h,h' \in \mathcal{H}_{|S}} |d_T(h.h')| = 
\sup_{h,h' \in \mathcal{H}_{|S}} \mathbb{E}_{x\sim T} \big[ \mathbb{I}(h(x) \neq h'(x)) \big]. 
\label{eq:max_discrep}
\end{equation}
This suggests the \textbf{MCD learning principle}: we need to minimise both the error on $S$ (so as to form the source-confined hypothesis space $\mathcal{H}_{|S}$) and the MCD loss on $T$. 
Note however that the last term $e^*$ is not considered in~\citep{saito2018,gpda} mainly due to the difficulty of estimating the target domain error. We will incorporate $e^*$ in our DG algorithm as described in the next section. 

We conclude the section by briefly reviewing how the MCD learning principle was exploited in previous works. In~\citep{saito2018} they explicitly introduce two classifier networks $h(x)=g(\phi(x))$ and $h'(x)=g'(\phi(x))$, where the classification heads $g$, $g'$ and the feature extractor $\phi$ are {\em cooperatively} updated to minimise the error on $S$ (thus implicitly obtaining $\mathcal{H}_{|S}$), they are updated {\em adversarially} to maximise (minimise) the MCD loss on $T$ with respect to $g$ and $g'$ ($\phi$, respectively). In~\citep{gpda}, 
they build a Gaussian process (GP) classifier on the feature space $\phi(x)$, in which  $\mathcal{H}_{|S}$ is attained by GP posterior inference. Minimisation of the MCD term is then  accomplished by the maximum margin learning which essentially enforces minimal overlap between the two largest posterior modes. 
Note that \citep{saito2018}'s strategy 
requires adversarial optimisation, 
and hence it is less suitable for our DG algorithm which will require adversarial generator learning: Having two adversarial learning components would make the training difficult since we need to find two nested equilibrium (saddle) points. We instead adopt the Bayesian hypothesis modeling approach of~\citep{gpda}. In the next section, we describe our approach in greater detail.

\section{\ourModelLong{} (\ourModel{})}\label{sec:main}
\vspace{-0.5em}
\noindent\textbf{Defining and optimising a hypothesis space.}\quad 
Our DG approach  aims to minimise the MCD loss, $\textrm{MCD}(\mathcal{H}_{|S};T)$ defined in (\ref{eq:max_discrep}). The first challenge is that the target domain data $T$ is not available in DG. Before we address it, we clarify the optimisation problem (i.e., what is the MCD loss optimised for?) and how the hypothesis spaces ($\mathcal{H}$ and $\mathcal{H}_{|S}$) are represented. 
The MCD loss is a function of {\em hypothesis space} $\mathcal{H}$ (or $\mathcal{H}_{|S}$), not a function of individual classifier $h$ in it. Hence, minimising the MCD loss amounts to choosing the best hypothesis space $\mathcal{H}$. To this end, we need to parametrise the hypothesis space (so as to frame it as a continuous optimisation), and our choice is the {\em Bayesian linear classifier with deterministic feature extractor}.  

We consider the conventional neural-network feed-forward classifier modeling: we have the feature extractor network $\phi_\theta(x)\in\mathbb{R}^d$ (with the weight parameters $\theta$) followed by the linear classification head $W=[w_1,\dots,w_C]$ ($C$-way classification, each $w_j\in\mathbb{R}^d$), where the class prediction is done by the softmax likelihood:
\begin{align}
P(y=j|x,\theta,W) \propto e^{w_j^\top \phi_\theta(x)}, \ \ \ \ j=1,\dots,C.
\label{eq:classifier_softmax}
\end{align}
So each configuration $(\theta,W)$ specifies a particular classifier $h$. To parametrise the hypothesis space $\mathcal{H}$ ($\ni h$), ideally we can consider {\em a parametric family of distributions} over $(\theta,W)$. Each distribution $P_\beta(\theta,W)$ specified by the parameter $\beta$ corresponds to a particular hypothesis space $\mathcal{H}$, and each sample $(\theta,W)\sim P_\beta(\theta,W)$ corresponds to a particular classifier $h\in\mathcal{H}$. Although this is conceptually simple, to have a tractable model in practice, we define $\theta$ to be {\em deterministic} parameters and only $W$ to be stochastic. A reasonable choice for $P(W)$, without any prior knowledge, is the standard Gaussian, $P(W) = \prod_{j=1}^C \mathcal{N}(w_j; 0, I)$. 

Now, 
we can represent a hypothesis space as $\mathcal{H} = \{ P(y|x,\theta,W) \ \vert \ W\sim P(W) \}$. Thus $\mathcal{H}$ is parametrised by $\theta$, and with $\theta$ fixed ($\mathcal{H}$ fixed), each sample $W$ from $P(W)$ instantiates a classifier $h\in\mathcal{H}$. 
The main benefit of this Bayesian hypothesis space modeling is that we can induce the {\em source-confined hypothesis space} $\mathcal{H}_{|S}$ (i.e., the set of classifiers that perform well on the source domain) in a principled manner by the posterior,  
\begin{equation}
P(W|S,\theta) \propto P(W) \cdot \prod_{(x,y)\sim S} P(y|x,\theta,W).
\label{eq:posterior}
\end{equation}
The posterior places most of its probability density on those samples (classifiers) $W$ that attain high likelihood scores on $S$ (under given $\theta$) while being smooth due to the prior. To ensure that the source domain $S$ is indeed explained well by the model, we further impose high data likelihood on $S$ as constraints for $\theta$,
\begin{align}
\theta \in \Theta_S \ \ \textrm{where} \ \Theta_S := \{\theta \ \vert \ \log P(S|\theta) \geq L_{th} \},
\label{eq:theta_constr}
\end{align}
where $L_{th}$ is the (constant) threshold that guarantees sufficient fidelity of the model to explaining $S$. 
Then it is reasonable to represent $\mathcal{H}_{|S}$ by the support of $P(W|S,\theta)$ for $\theta\in\Theta_S$, postulating that $\mathcal{H}_{|S}$ exclusively contains smooth classifiers $h$ that perform well on $S$. Formally, the source-confined hypothesis space can be parametrised as: 
\begin{align}
\mathcal{H}_{|S}(\theta) = \{ P(y|x,\theta,W) \ \vert \ W\sim P(W|S,\theta) \} \ \  \textrm{for} \ \theta\in\Theta_S,
\label{eq:S_confined_H}
\end{align}
where we use the notation $\mathcal{H}_{|S}(\theta)$ to emphasise its dependency on $\theta$. 
Intuitively, the hypothesis space $\mathcal{H}_{|S}$ is identified by choosing the feature space (i.e., choosing $\theta\in\Theta_S$), and individual classifiers $h\in\mathcal{H}_{|S}$ are realised by the Bayesian posterior samples $W\sim P(W|S,\theta)$ (inferred on the chosen feature space). 
Since the posterior $P(W|S,\theta)$ in (\ref{eq:S_confined_H}) and the marginal likelihood $\log P(S|\theta)$ in (\ref{eq:theta_constr}) do not admit closed forms in general, we adopt the variational inference technique to approximate them. We defer the detailed derivations (Sec.~\ref{sec:main_concrete}) for now, and return to the MCD minimisation problem since we have defined the hypothesis space representation.


\vspace{0.2cm}\noindent\textbf{Optimising a worst-case target domain.}\quad 
For the DG problem, we cannot directly apply the MCD learning principle since the target domain $T$ is unknown during the training stage. Our key idea is to consider the worst-case scenario where the target domain $T$  maximises the MCD loss. 
This naturally forms minimax-type optimisation,
\begin{align}
\min_{\theta\in\Theta_S} \max_T \ \textrm{MCD}(\mathcal{H}_{|S}(\theta); T). 
\label{eq:minimax}
\end{align}
To solve the saddle-point optimisation (\ref{eq:minimax}), we adopt the adversarial learning strategy with a generator network~\citep{gan14}. The generator for $T$ has to synthesise samples $x$ of $T$ that need to satisfy three conditions: (\textbf{C1}) The generated samples maximally baffle the classifiers in $\mathcal{H}_{|S}$ to have least consensus in prediction (for inner maximisation); (\textbf{C2}) $T$ still retains the same semantic class information as the source domain $S$ (for the definition of DG); and (\textbf{C3}) The generated samples in $T$ need to be distinguishable along their classes\footnote{This condition naturally originates from the solvability of the DG problem.}. 

\vspace{0.2cm}\noindent\textbf{Paramaterising domains.}\quad
To meet these conditions, we generate target domain images using  Fourier frequency spectra. We specifically build a generator network that synthesises {\em amplitude} images in the Fourier frequency domain. The synthesised amplitude images are then combined with the {\em phase} images sampled from the source domain $S$ to construct new samples $x\in T$ by inverse Fourier transform. 
This is motivated by signal processing where it is widely believed that the frequency phase spectra capture the semantic information of signals, while the  amplitudes take charge of non-semantic (e.g., style) aspects of the  signals \citep{oppenheim1981importance}. Denoting the amplitude generator network as $G_\nu(\epsilon)$ with parameters $\nu$ and random noise input  $\epsilon\sim\mathcal{N}(0,I)$, our target sampler $(x,y)\sim T$ are generated 
as follows:
\begin{enumerate}
\item $(x_S, y_S) \sim S$ \ \ \ \ \ \ \ \ \ \ \ \ \ \ \ \ \ \ \ \ \ \ \ \ \ \ \ \ \ \ \ (Sample an image and its class label from $S$)
\item $A_S \angle P_S = \mathcal{F}(x_S)$ \ \ \ \ \ \ (Fourier transform to have amplitude and phase for $x_S$)
\item $A = G_\nu(\epsilon)$, $\epsilon\sim\mathcal{N}(0,I)$ \ \ \ \ \ \ \ \ \ \ \ \ \ \ \ \ \ \ \ \ \ \ (Generate an amplitude image from $G$)
\item $x = \mathcal{F}^{-1}(A\angle P_S)$, $y=y_S$ \ \ \ \ \ \ \ \ (Construct target data with the synthesised $A$)
\end{enumerate}
Here, $\mathcal{F}(\cdot)$ is the 2D Fourier transform, $F(u,v) = \mathcal{F}(x) = \iint x(h,w) e^{-i(hu+wv)}dh dw$, 
and $A \angle P$ stands for the polar representation of the Fourier frequency responses (complex numbers) for the amplitude image $A$ and the phase image $P$. That is, $A \angle P = A \cdot e^{i\cdot P} = A \cdot (\cos P + i \sin P)$ with $i=\sqrt{-1}$, where all operations are element/pixel-wise. 
Note that 
we set $y=y_S$ in step 4 since the original phase (semantic) information $P_S$ is retained in the synthesised $x$. 

\vspace{0.2cm}\noindent\textbf{Algorithm summary.}\quad
Finally the worst-case target MCD learning can be solved by adversarial learning, which can be implemented as an alternating optimisation: 
\begin{align}
&\textrm{(Fix $\nu$)} \ \ \min_{\theta\in\Theta_S} \  \textrm{MCD}(\mathcal{H}_{|S}(\theta); T(\nu)) \label{eq:general_alt1}\\
&\textrm{(Fix $\theta$)} \ \ \ \max_\nu \  \textrm{MCD}(\mathcal{H}_{|S}(\theta); T(\nu))
\label{eq:general_alt2}
\end{align}
We used $T(\nu)$ to emphasise functional dependency of target images on the generator parameters $\nu$. 
Note that although the MCD loss in DA 
can be computed {\em without} the target domain labels (recall the definition (\ref{eq:max_discrep})), in our DG case the class labels for the generated target data are available, as induced from the phase $P_S$ (i.e., $y=y_S$ in step 4). Thus we can modify the MCD loss by incorporating the target class labels. 
In the following we provide concrete derivations using the variational posterior inference, and propose a modified MCD loss that takes into account the induced target class labels.

\subsection{Concrete Derivations using Variational Inference}\label{sec:main_concrete}
\vspace{-0.5em}
\noindent\textbf{Source-confined hypothesis space by variational inference.}\quad
The posterior $P(W|S,\theta)$ does not admit a closed form, and 
we approximate 
$P(W|S,\theta)$ by the Gaussian variational density,
\begin{equation}
Q_\lambda(W) = \prod_{j=1}^C \mathcal{N}(w_j; m_j, V_j),
\label{eq:inf_q}
\end{equation}
where $\lambda:=\{m_j,V_j\}_{j=1}^C$ constitutes the variational parameters. 
To enforce $Q_\lambda(W) \approx P(W|S,\theta)$, we optimise the evidence lower bound (ELBO), 
\begin{equation}
\textrm{ELBO}(\lambda,\theta;S) :=
  \sum_{(x,y)\sim S} \mathbb{E}_{Q_\lambda(W)} \big[ 
    \log P(y|x,W,\theta) \big]  - 
  \textrm{KL} \big( Q_\lambda(W) || P(W) \big),
\label{eq:elbo}
\end{equation}
which is the lower bound of the marginal data likelihood $\log P(S|\theta)$ (Appendix~\ref{sec:derivations} for derivations). Hence maximising $\textrm{ELBO}(\lambda,\theta;S)$ with respect to $\lambda$ tightens the posterior approximation $Q_\lambda(W) \approx P(W|S,\theta)$, while maximising it with respect to $\theta$ leads to high data likelihood $\log P(S|\theta)$. The latter has the very effect of imposing the constraints $\theta\in\Theta_S$ in (\ref{eq:general_alt1}) since one can transform constrained optimisation into a regularised (Lagrangian) form equivalently~\citep{cvx}. 



\vspace{0.2cm}\noindent\textbf{Optimising the MCD loss.}\quad
The next thing is to minimise the MCD loss, $\textrm{MCD}(\mathcal{H}_{|S}(\theta); T)$ with the current target domain $T$ generated by the generator network $G_\nu$. That is, solving (\ref{eq:general_alt1}). We  follow the maximum margin learning strategy from~\citep{gpda}, where the idea is to enforce the prediction consistency for different classifiers (i.e., posterior samples) $W\sim Q_\lambda(W)$ on $x\sim T$ by separating the highest class score from the second highest by large margin. To understand the idea, let $j^*$ be the model's predicted class label for $x\sim T$, or equivalently let $j^*$ have the highest class score $j^* = \arg\max_j w_j^\top \phi(x)$ as per (\ref{eq:classifier_softmax}). (We drop the subscript in $\phi_\theta(x)$ for simplicity in notation.)
We let $j^\dagger$ be the second most probable class, i.e., $j^\dagger = \arg\max_{j\neq j^*} w_j^\top \phi(x)$. Our model's class prediction would change if $w_{j^*}^\top \phi(x) < w_{j^\dagger}^\top \phi(x)$ for some $W\sim Q_\lambda(W)$, which leads to {\em discrepancy} of classifiers. To avoid such overtaking, we need to ensure that the (plausible) {\em minimal} value of $w_{j^*}^\top\phi(x)$ is greater than the (plausible) {\em maximal} value of $w_{j^\dagger}^\top\phi(x)$. Since the score (logit) $f_j(x) := w_j^\top\phi(x)$ is 
Gaussian under $Q_\lambda(W)$, namely
\begin{align}
f_j(x) \sim \mathcal{N}(\mu_j(x), \sigma_j(x)^2) \ \ \textrm{where} \ \ 
\mu_j(x) = m_j^\top\phi(x), \ \sigma_j^2(x)=\phi(x)^\top V_j \phi(x),
\label{eq:logit_gaussian}
\end{align}
the prediction consistency is achieved by enforcing: $\mu_{j^*}(x) -\alpha \sigma_{j^*}(x) > \mu_{j^\dagger}(x) + \alpha \sigma_{j^\dagger}(x)$,  where we can choose $\alpha=1.96$ for $2.5\%$ rare one-sided chance. 
By introducing slack variables $\xi(x) \geq 0$, 
\begin{equation}
\mu_{j^*}(x) - \alpha \sigma_{j^*}(x) 
\geq 1 + \max_{j\neq j^*} \big( \mu_j(x) + \alpha \sigma_j(x) \big) - \xi(x). 
\label{eq:post_sep_ineq}
\end{equation}
Satisfying the constraints amounts to fulfilling the desideratum of MCD minimisation, essentially imposing prediction consistency of classifiers.  Note that we add the constant $1$ in the right hand side of (\ref{eq:post_sep_ineq}) for the normalisation purpose to prevent the scale of $\mu$ and $\sigma$ from being arbitrary small. The constraints in (\ref{eq:post_sep_ineq}) can be translated into the following MCD loss (as a function of $\theta$):
\begin{align}
\textrm{MCD}(\theta; T) 
:= \mathbb{E}_{x\sim T} 
\Big( 1 + \mathcal{T}^2 \big( \mu_j(x) + \alpha \sigma_j(x) \big) - \mathcal{T}^1 \big( \mu_j(x) - \alpha \sigma_j(x) \big) \Big)_+
\label{eq:max_marg_obj}
\end{align}
where $\mathcal{T}^k$ is the operator that selects the top-$k$ element, 
and $(a)_+ = \max(0,a)$.

\begin{figure}[t!]
\vspace{-3.0em}
\begin{center}
%
\centering
\includegraphics[trim = 4mm 4.5mm 4mm 4mm, clip, scale=0.242]{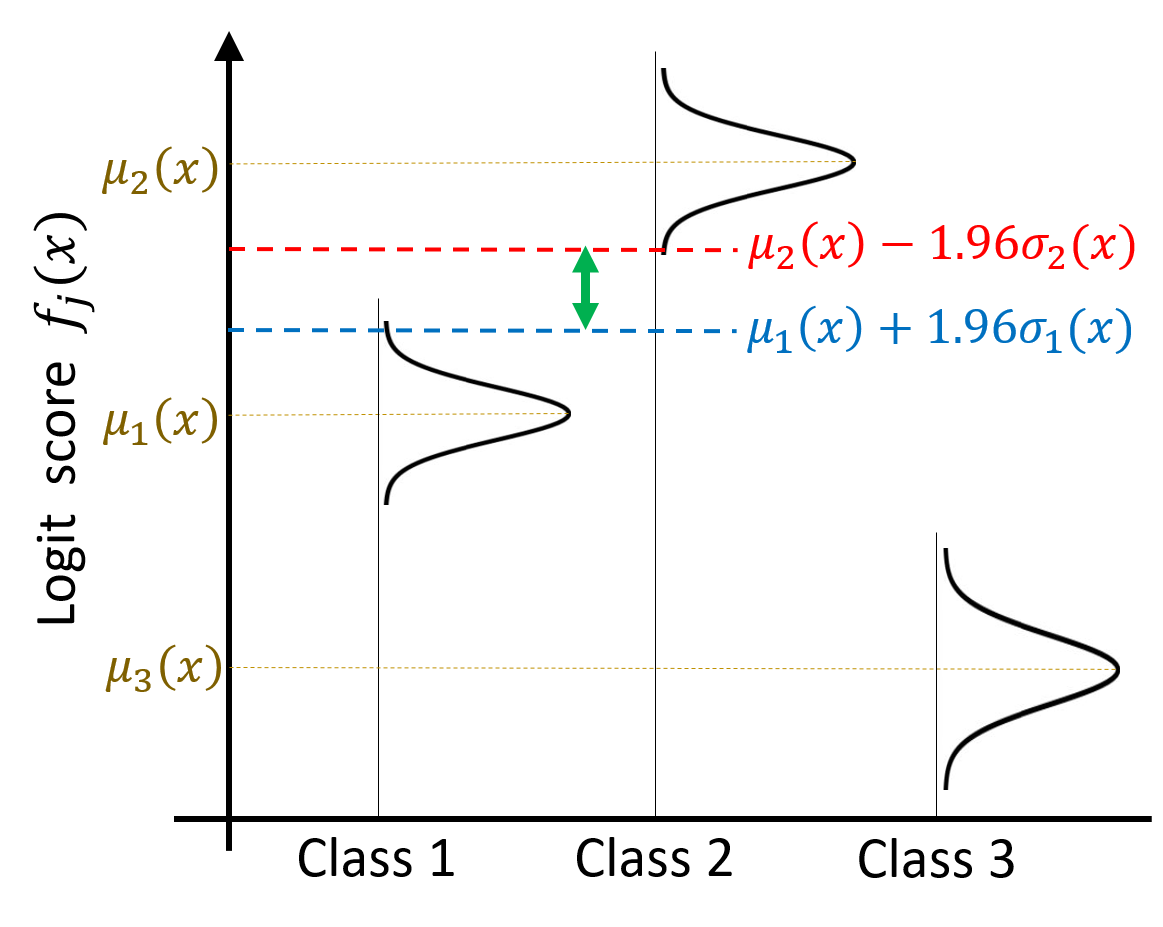} \ \ 
\includegraphics[trim = 18mm 4.5mm 4mm 4mm, clip, scale=0.242]{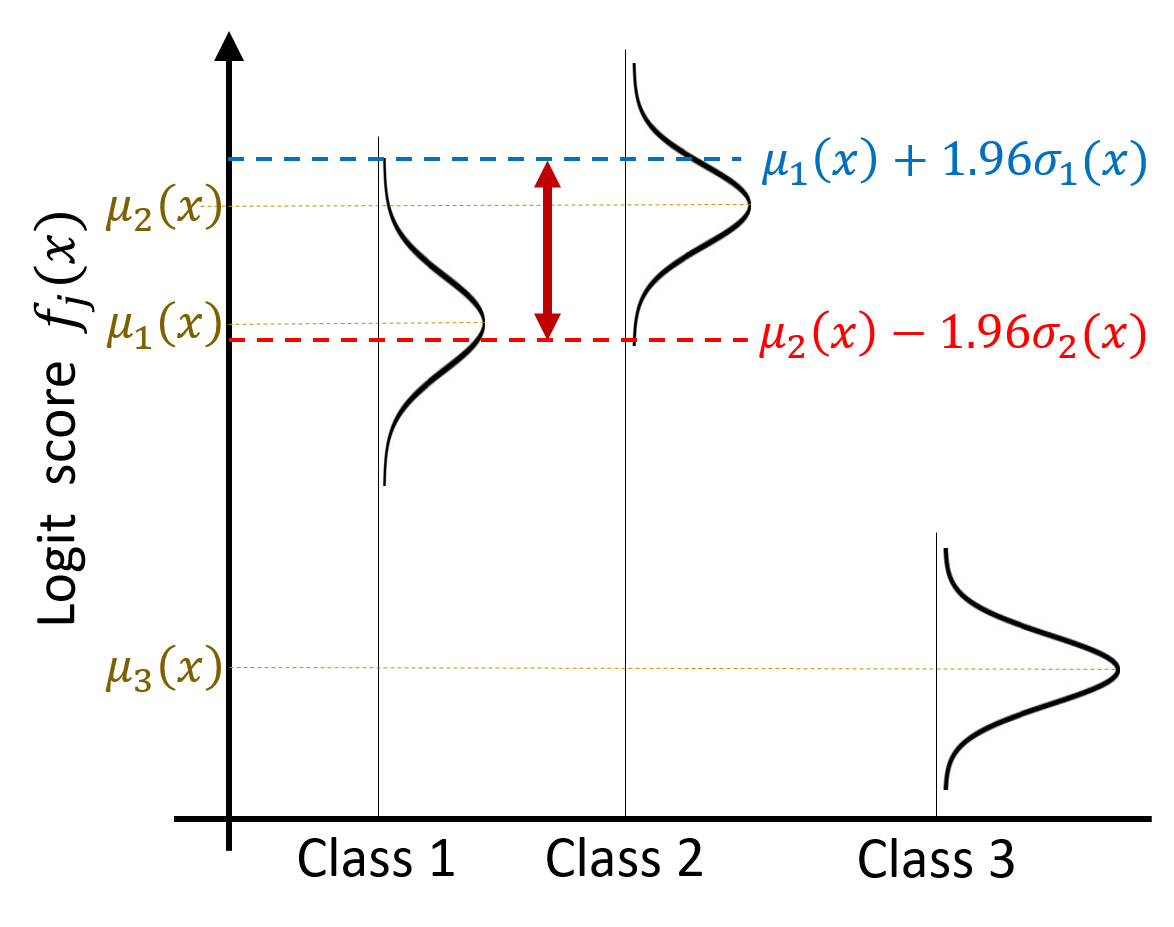} \ \ 
\includegraphics[trim = 18mm 4.5mm 4mm 4mm, clip, scale=0.242]{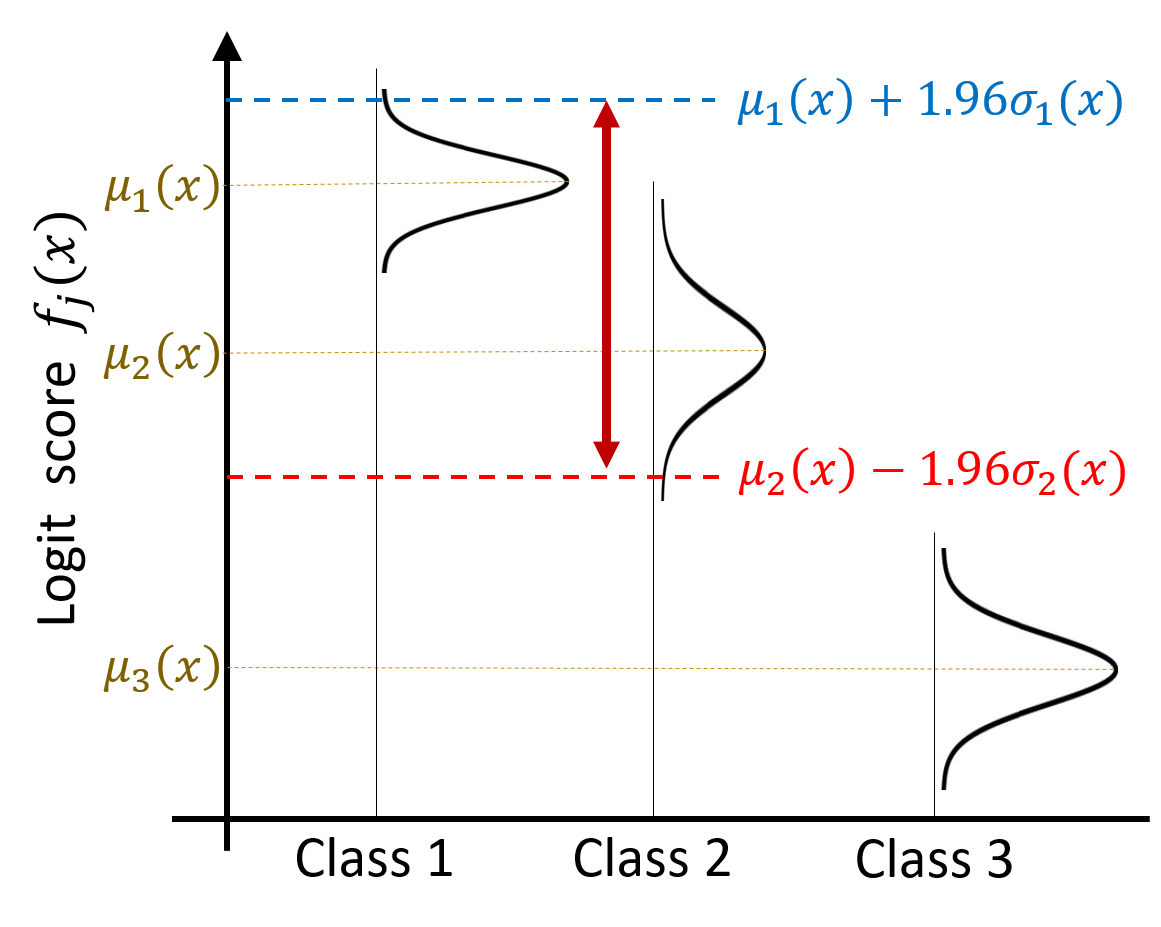}
\end{center}
\vspace{-1.5em}
\caption{Illustration of the SMCD loss on three different hypothesis spaces $\mathcal{H}_{|S}$ shown in three panels. For $C=3$-way classification case, each panel shows the class logit scores (Gaussian random) $f_j(x)\sim \mathcal{N}(\mu_j(x),\sigma_j(x)^2)$ for $j=1,2,3$, at some input $x\in T$. We assume that the true (induced) class label $y=2$.
(\textbf{Left}) Since the mean logit for class 2, $\mu_2(x)$ is the maximum among others, the prediction is {\em marginally} correct (from softmax). Beyond that, the logit of the worst plausible hypothesis for class 2, $\mu_2(x)-1.96 \sigma_2(x)$ is greater than that of the runner-up class 1, $\mu_1(x)+1.96 \sigma_1(x)$ by some positive margin (green arrow), meaning there is little chance of prediction overtaking (so, consistent); equivalently, the SMCD loss is small. (\textbf{Middle}) Prediction is marginally correct, but prediction overtaking is plausible,  indicated by the negative margin (red arrow); the SMCD loss is large. (\textbf{Right}) Incorrect marginal prediction (to class 1) with more severe negative margin (red arrow); the SMCD loss is even larger. 
}
\vspace{-0.5em}
\label{fig:logit_cases}
\end{figure}

\vspace{0.2cm}\noindent\textbf{Modified MCD loss.}\quad
The above MCD loss does not utilise the target domain class labels $y=y_S$ that are induced from the phase information $P_S$ (Recall the target domain data generation steps $1\sim 4$ above). To incorporate the supervised data $\{(x,y)\} \in T$ in the generated target domain, we modify the MCD loss as follows: 
First, instead of separating the margin between the two largest logit scores as in the MCD, we maximise the margin between the logit for the given class $y$ and the largest logit among the classes other than $y$. That is, we replace the constraints (\ref{eq:post_sep_ineq}) with the following:
\begin{equation}
\mu_{y}(x) - \alpha \sigma_{y}(x) 
\geq 1 + \max_{j\neq y} \big( \mu_j(x) + \alpha \sigma_j(x) \big) - \xi(x),
\label{eq:modified_post_sep_ineq}
\end{equation}
where $y$ is the class label (induced from the phase information) for the generated instance $x$. See Fig.~\ref{fig:logit_cases} for illustration of the idea. 
Consequently, our new MCD loss (coined {\em supervised} MCD or $\textrm{SMCD}$ for short) is defined as follows:
\begin{align}
\textrm{SMCD}(\theta; T) 
:= \mathbb{E}_{(x,y)\sim T} 
\Big( 1 + \max_{j\neq y} \big( \mu_j(x) + \alpha \sigma_j(x) \big) - \big( \mu_y(x) - \alpha \sigma_y(x) \big) \Big)_+.
\label{eq:modified_max_marg_obj}
\end{align}
Here the variational parameters $\lambda$ is treated as constant 
since the only role of $\lambda$ is to maximise the ELBO. It should be noted that  (\ref{eq:modified_max_marg_obj}) essentially aims at maximising the logit for the given class $y$ (the last term), or equivalently, classification error minimisation on $T$, and at the same time minimising the logit for the runner-up class (the middle max term). 
Surprisingly, the former amounts to minimising the minimal source-target error term $e^*(\mathcal{H};S,T)$ in the generalisation bound (\ref{eq:thm_2010}), which we have left out so far. That is, $e^*(\mathcal{H};S,T) = \min_{h\in\mathcal{H}} e_S(h) + e_T(h) \approx \min_{h\in\mathcal{H}_{|S}} e_T(h)$, 
and the last term of the SMCD loss leads to $\theta$ that makes $e_T(h)$ small for all $h\in\mathcal{H}_{|S}(\theta)$. 
Moreover, 
minimising the logit for the runner-up class (the middle max term of the SMCD) has the effect of margin maximisation. 

\vspace{-0.7em}
\vspace{0.2cm}\noindent\textbf{Algorithm summary.}\quad 
Our \ourModel{} algorithm can be understood as  {\em MCD-based DA with adversarial amplitude generated target domain}. It entails the following alternating optimisation ($\eta>0$ is the trade-off hyperparameter for SMCD):
\begin{enumerate}
%
%
\item $\min_{\lambda,\theta} \ 
-\textrm{ELBO}(\lambda,\theta;S) +\eta \textrm{SMCD}(\theta;T)$ \ \ \ \ \ \ (model learning $+$ VI; $\nu$ fixed)
\item $\max_\nu \ \textrm{SMCD}(\theta;T)$ \ \ \ \ \ \ \ \ \ \ \ \ \ \ \ \ \ \ \ \ \ \ \ \ (adversarial generator learning; $\theta$, $\lambda$ fixed)
%
\end{enumerate}
Our algorithm 
is summarised in Alg.~\ref{alg:main_swad} (in Appendix) and illustrated schematically in Fig.~\ref{fig:main}. At test time, we can apply the classifier (\ref{eq:classifier_softmax}) with the learned $\theta$ and any sample $W\sim Q_\lambda(W)$ to target domain inputs to predict class labels. In our experiments, we take the posterior means $w_j=m_j$ instead of sampling from $Q_\lambda(W)$. 

\subsection{Further Considerations} 
\vspace{-0.5em}

\textbf{Post-synthesis mixup of generated amplitude images.}
\quad 
In our adversarial learning, the amplitude generator network $G_\nu$ synthesises target domain image samples that have highly challenging amplitude spectra to the current model. Although we retain the phase information from source domains, unconstrained amplitude images can potentially alter the semantic content destructively (e.g., a constant zero amplitude image would zero out the image content), rendering it impossible to classify.
To this end, instead of using the generator's output $A=G_\nu(\epsilon)$ directly, we combine it with the source domain amplitude image corresponding to the phase image by simple mixup. That is, by letting $A_S$ be the amplitude spectra corresponding to the phase $P_S$, we alter $A$ as:
\begin{align}
A \leftarrow \lambda A + (1-\lambda) A_S \ \ \textrm{where} \ \ \lambda\sim\textrm{Uniform}(0,\alpha).
\label{eq:post_mixup}
\vspace{-0.5em}
\end{align}
This post-synthesis mixup can address our desideratum \textbf{C3} that we discussed before, that is, the generated samples for the target domain need to be distinguishable by class to solve the DG problem. Post-synthesis mixup, ensures synthesised amplitude images lie closer to the amplitude manifold of the source data, ensuring the model can solve the classification problem.

\noindent\textbf{Dense model averaging (SWAD).}\quad
We found that the DG training becomes more stable and the target-domain test performance becomes more consistent when we use the dense model averaging strategy SWAD~\citep{swad}. We adopt the SWAD model averaging for the variational and model parameters $(\lambda,\theta)$ while the generator network is not averaged. 

\noindent\textbf{Amplitude image structures.}\quad
From the definition of the Fourier transform, the frequency domain function should be  even-conjugate, i.e., $F(-u,-v) = \overline{F(u,v)}$, for the real-valued images.  
This implies that amplitude images are symmetric. Conversely, if the amplitude images are symmetric, inverse Fourier transform returns real-valued signals. Thus when generating amplitude images, we only generate the non-redundant part (frequencies) of the amplitude images. Also, the amplitude should be non-negative. We keep these constraints in mind when designing the generator network.
\section{Related Work}\label{sec:related}
\vspace{-0.5em}

\noindent\textbf{
MCD.}\quad Several studies have used the MCD principle for domain adaptation, to align a source model to unlabeled target data \citep{saito2018,gpda,lu2020stochastic}. We uniquely exploit the MCD principle for the DG problem, in the absence of target data, by using MCD to synthesise worst-case target domain data, as well as to adapt the model to that synthesised domain.

\noindent\textbf{Augmentation approaches to DG.}\quad Several DG approaches have been proposed based on data augmentation. Existing approaches either define augmentation heuristics \citep{mixstyle,fact}, or exploit \emph{domain} adversarial learning -- i.e., confusing a \emph{domain classifier} \citep{shankar2018crossgrad,zhou2020deepDG}. Our adversarial learning is based on the much stronger (S)MCD principle that confuses a category classifier. This provides much harder examples for robust learning, while our Fourier amplitude synthesis ensures the examples are actually recognisable.

\noindent\textbf{Alignment approaches to DG.}\quad Several approaches to DG are based on aligning between multiple source domains \citep{coral,dann,cdann,mmd}, under the assumption that a common feature across all source domains will be good for a held out target domain. Differently, we use the MCD principle to robustify our source trained model by aligning it with the synthesised worst-case target domain.

\section{Experiments}\label{sec:expmt}
\vspace{-0.5em}
We 
test our approach on the DomainBed benchmark~\citep{domainbed},  
including: 
\textbf{PACS}~\citep{pacs}, 
\textbf{VLCS}~\citep{vlcs}, 
\textbf{OfficeHome}~\citep{office_home}, 
\textbf{TerraIncognita}~\citep{ti}, 
and 
\textbf{DomainNet}~\citep{domain_net}. 
For each dataset, we adopt the standard leave-one-domain-out source/target domain splits. The overall training/test protocols are similar to~\citep{domainbed,swad}.  
We use the ResNet-50~\citep{resnet} as our feature extractor backbone, which is initialised by the pretrained weights on ImageNet~\citep{deng2009imagenet}. For the generator network, we found that a linear model performed the best for the noise dimension 100. Our model is trained by the Adam optimiser~\citep{adam} on machines with single Tesla V100 GPUs. 
The hyperparameters introduced in our model (e.g., SMCD trade-off $\eta$) and the general ones (e.g., learning rate, SWAD regime hyperparameters, maximum numbers of iterations) are chosen by grid search on the validation set according to the DomainBed protocol \citep{domainbed}. For instance, $\eta=0.1$ for all datasets. 
The implementation details including chosen hyperparameters can be found in Appendix~\ref{sec:impl_details}.


\subsection{Main Results}\label{sec:expmt_main}
\vspace{-0.5em}

\definecolor{lor}{rgb}{1,0.85,0}
\definecolor{or}{rgb}{1,0.60,0}
\definecolor{dor}{rgb}{1,0.20,0}
\begin{table}[t!]
\vspace{-4.5em}
\centering
\caption{Average accuracies on DomainBed datasets. 
Note: $^\dagger$ indicates that the results are excerpted from the published papers or~\citep{domainbed}. Our own runs are reported without $^\dagger$. 
Note that FACT~\citep{fact} adopted a slightly different data/domain split protocol from DomainBed's, explaining discrepancy on PACS. 
}
\vspace{-0.0em}
\label{tab:main}
\begin{scriptsize}
\centering
\scalebox{0.95}{
\begin{tabular}{lccccc|c}
\toprule
Algorithm \ \ \ \ \ \ \ \ & \ \ PACS \ \ & \ VLCS \ & OfficeHome & TerraInc. & DomainNet & \ \ Avg. \ \ \\
\midrule
ERM~\citep{swad}$^\dagger$\Tstrut & $84.2$ & $77.3$ & $67.6$ & $47.8$ & $44.0$ & $64.2$ \\ 
IRM~\citep{irm}$^\dagger$\Tstrut & $83.5$ & $78.6$ & $64.3$ & $47.6$ & $33.9$ & $61.6$ \\ 
GroupDRO~\citep{group_dro}$^\dagger$\Tstrut & $84.4$ & $76.7$ & $66.0$ & $43.2$ & $33.3$ & $60.7$ \\ 
I-Mixup~\citep{imixup_1,imixup_2,imixup_3}$^\dagger$\Tstrut & $84.6$ & $77.4$ & $68.1$ & $47.9$ & $39.2$ & $63.4$ \\ 
MLDG~\citep{mldg}$^\dagger$\Tstrut & $84.9$ & $77.2$ & $66.8$ & $47.8$ & $41.2$ & $63.6$ \\ 
CORAL~\citep{coral}$^\dagger$\Tstrut & $86.2$ & $78.8$ & $68.7$ & $47.7$ & $41.5$ & $64.5$ \\ 
MMD~\citep{mmd}$^\dagger$\Tstrut & $84.7$ & $77.5$ & $66.4$ & $42.2$ & $23.4$ & $58.8$ \\ 
DANN~\citep{dann}$^\dagger$\Tstrut & $83.7$ & $78.6$ & $65.9$ & $46.7$ & $38.3$ & $62.6$ \\ 
CDANN~\citep{cdann}$^\dagger$\Tstrut & $82.6$ & $77.5$ & $65.7$ & $45.8$ & $38.3$ & $62.0$ \\ 
MTL~\citep{mtl}$^\dagger$\Tstrut & $84.6$ & $77.2$ & $66.4$ & $45.6$ & $40.6$ & $62.9$ \\ 
SagNet~\citep{sagnet}$^\dagger$\Tstrut & $86.3$ & $77.8$ & $68.1$ & $48.6$ & $40.3$ & $64.2$ \\
ARM~\citep{arm}$^\dagger$\Tstrut & $85.1$ & $77.6$ & $64.8$ & $45.5$ & $35.5$ & $61.7$ \\
VREx~\citep{vrex}$^\dagger$\Tstrut & $84.9$ & $78.3$ & $66.4$ & $46.4$ & $33.6$ & $61.9$ \\
RSC~\citep{rsc}$^\dagger$\Tstrut & $85.2$ & $77.1$ & $65.5$ & $46.6$ & $38.9$ & $62.7$ \\
Mixstyle~\citep{mixstyle}$^\dagger$\Tstrut & $85.2$ & $77.9$ & $60.4$ & $44.0$ & $34.0$ & $60.3$ \\
FACT~\citep{fact}$^\dagger$\Tstrut & $88.2$ & $-$ & $66.6$ & $-$ & $-$ & $-$ \\ 
FACT~\citep{fact}\Tstrut & $86.4$ & $76.6$ & $66.6$ & $45.4$ & $42.6$ & $63.5$ \\ 
Amp-Mixup~\citep{fact}\Tstrut & $84.7$ & $75.9$ & $64.0$ & $46.8$ & $42.0$ & $62.7$ \\ 
\hline
SWAD~\citep{swad}$^\dagger$\Tstrut & $88.1$ & $79.1$ & $70.6$ & $50.0$ & $46.5$ & $66.9$ \\ 
FACT$+$SWAD\Tstrut & $88.1$ & $77.7$ & $70.6$ & $51.0$ & $46.7$ & $66.8$ \\ 
Amp-Mixup$+$SWAD\Tstrut & $88.1$ & $78.2$ & $70.3$ & $51.2$ & $46.4$ & $66.8$ \\ 
\hline
(Proposed) \ourModel{}\Tstrut & ${\bf 89.3}$ & ${\bf 79.5}$ & ${\bf 71.5}$ & ${\bf 52.4}$ & ${\bf 47.1}$ & ${\bf 68.0}$ \\
\bottomrule
\end{tabular}
}
\end{scriptsize}
\vspace{-1.0em}
\end{table}


The test accuracies averaged over target domains are summarised in Table~\ref{tab:main}, where the results for individual target domains are reported in Appendix~\ref{sec:full_results}. 
The proposed approach performs the best for all datasets among the competitors, and the difference from the second best model (SWAD) is significant (about $1.1\%$ margin). 
We particularly contrast with two recent approaches: \textbf{SWAD}~\citep{swad} that adopts the dense model averaging with the simple ERM loss and \textbf{FACT}~\citep{fact} that uses the Fourier amplitude mixup as means of data augmentation with additional student-teacher regularisation. 
%

First, SWAD~\citep{swad} is the second best model in Table~\ref{tab:main}, implying that the simple ERM loss combined with the dense model averaging that seeks for flat minima is quite effective, also observed previously~\citep{domainbed}. 
FACT~\citep{fact} utilises the Fourier amplitude spectra similar to our approach, but their main focus is {\em data augmentation}, producing more training images by amplitude mixup of source domain images. FACT also adopted the so-called {\em teacher co-regularisation} which forces the orders of the class prediction logits to be consistent between teacher and student models on the amplitude-mixup data. To disentangle the impact of these two components in FACT, we ran a model called \textbf{Amp-Mixup} that is simply FACT without teacher  co-regularisation. 
The teacher co-regularisation yields further improvement in the average accuracy (FACT $>$ Amp-Mixup in the last column of Table~\ref{tab:main}), verifying the claim in~\citep{fact}, although FACT is slightly worse than Amp-Mixup on VLCS and TerraIncognita.  

We also modified FACT and Amp-Mixup models by incorporating the SWAD model averaging (FACT$+$SWAD and Amp-Mixup$+$SWAD in the table). Clearly they perform even better in combination with  SWAD. Since Amp-Mixup$+$SWAD can be seen as dropping the teacher regularisation and adopting the SWAD (regularisation) strategy instead, we can say that SWAD is more effective regularisation than student-teacher. Nevertheless, despite the utilisation of amplitude-mixup augmentation, it appears that FACT and Amp-Mixup have little improvement over the ERM loss even when the SWAD strategy is used. This signifies the effect of the adversarial Fourier-based target domain generation in our approach which exhibits significant improvement over ERM and SWAD. 
%

\subsection{Further Analysis}\label{sec:ablation}
\vspace{-0.5em}

\begin{figure}[t!]
\vspace{-3.0em}
\begin{center}
%
\centering
\includegraphics[trim = 2mm 2mm 5mm 4mm, clip, scale=0.225
]{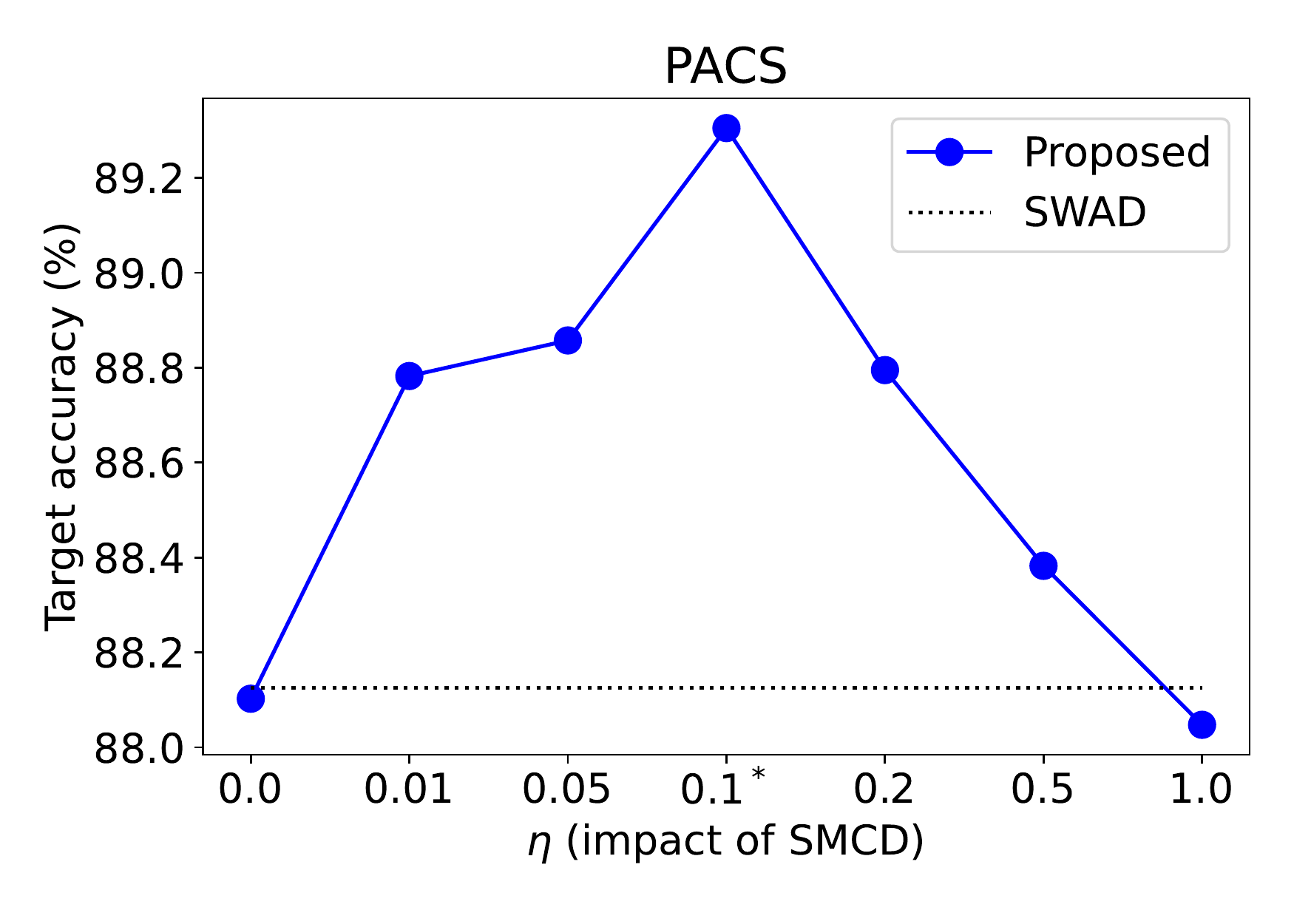}
\includegraphics[trim = 2mm 2mm 5mm 4mm, clip, scale=0.225
]{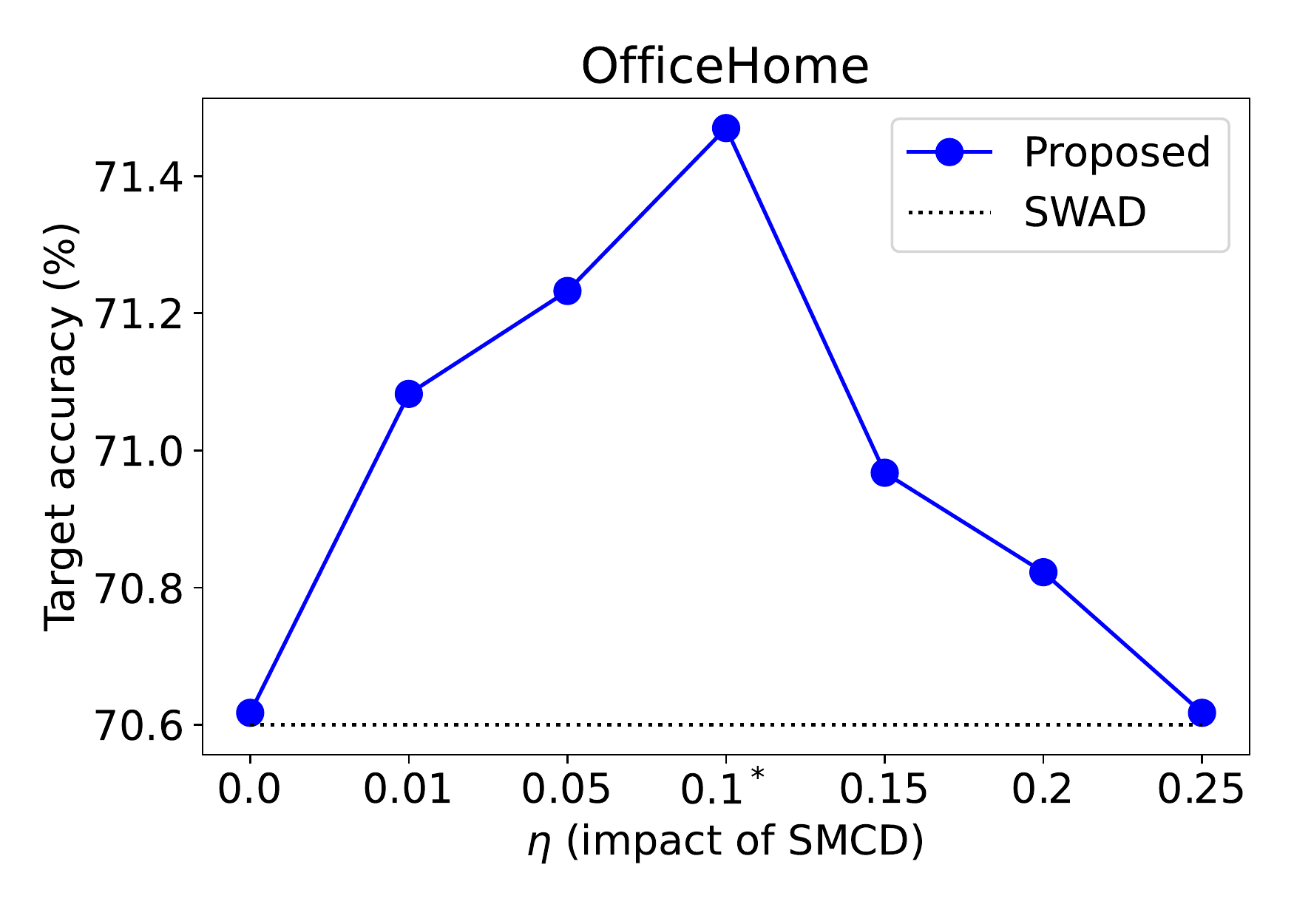}
\end{center}
\vspace{-2.0em}
\caption{Sensitivity to $\eta$ (SMCD trade-off) on PACS and OfficeHome. 
}
\label{fig:sensitivity_eta}
\end{figure}

\textbf{Sensitivity to $\eta$ (SMCD strength).}\quad 
We analyze sensitivity of the target domain generalisation performance to the SMCD trade-off hyperparameter $\eta$. 
We run our algorithm with different values of $\eta$. The results are shown in Fig.~\ref{fig:sensitivity_eta}. Note that $\eta=0$ ignores the SMCD loss term (thus generator has no influence on the model training), which corresponds to the ERM approach. 
The test accuracy of the proposed approach remains significantly better than ERM/SWAD for all those $\eta$ with moderate variations around the best value. 
See Appendix~\ref{sec:full_results} for the results on individual target domains. 

\begin{figure}[t!]
\vspace{-1.2em}
\begin{center}
%
\centering
\includegraphics[trim = 2mm 2mm 5mm 4mm, clip, scale=0.225
]{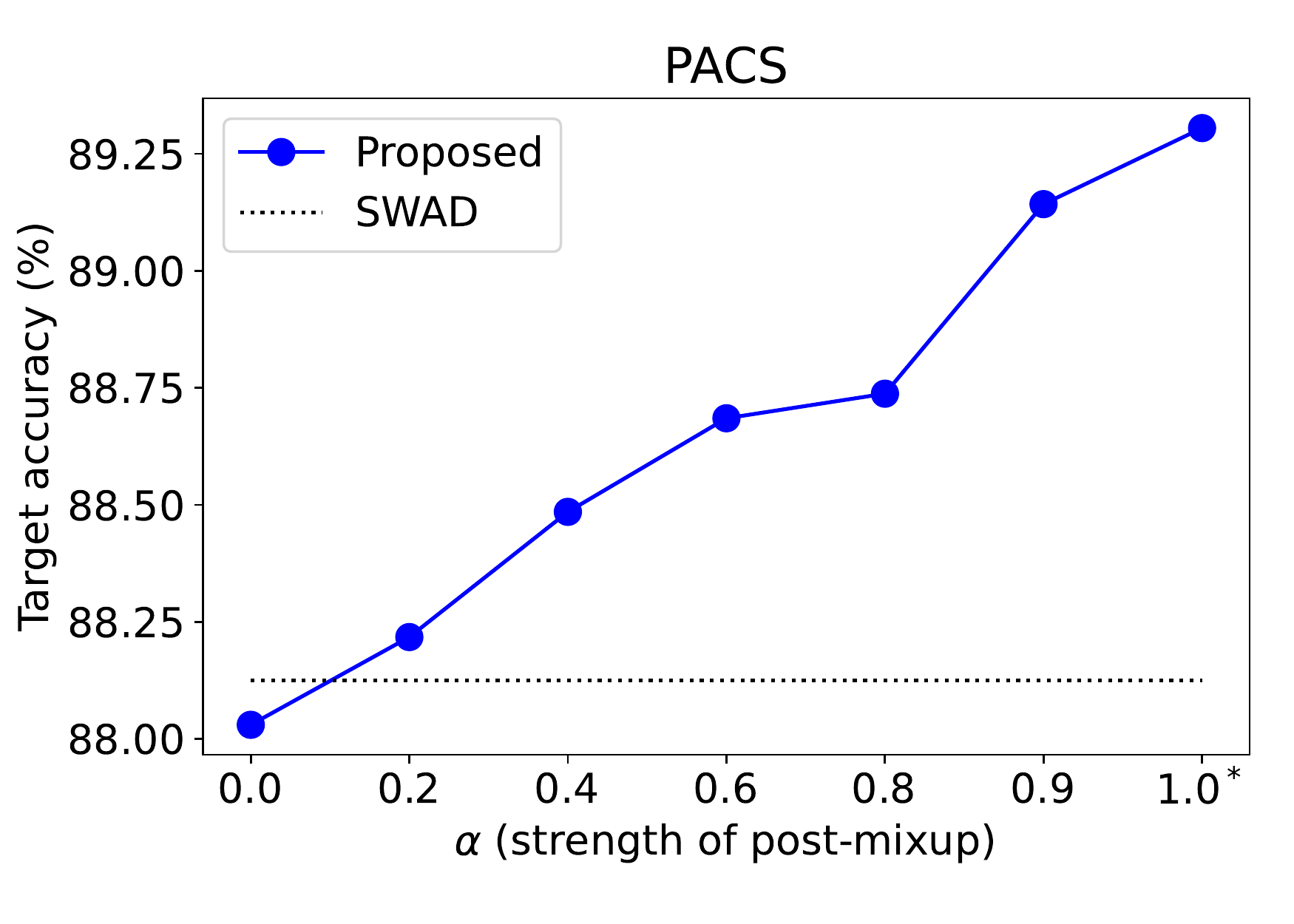} 
\includegraphics[trim = 2mm 2mm 5mm 4mm, clip, scale=0.225
]{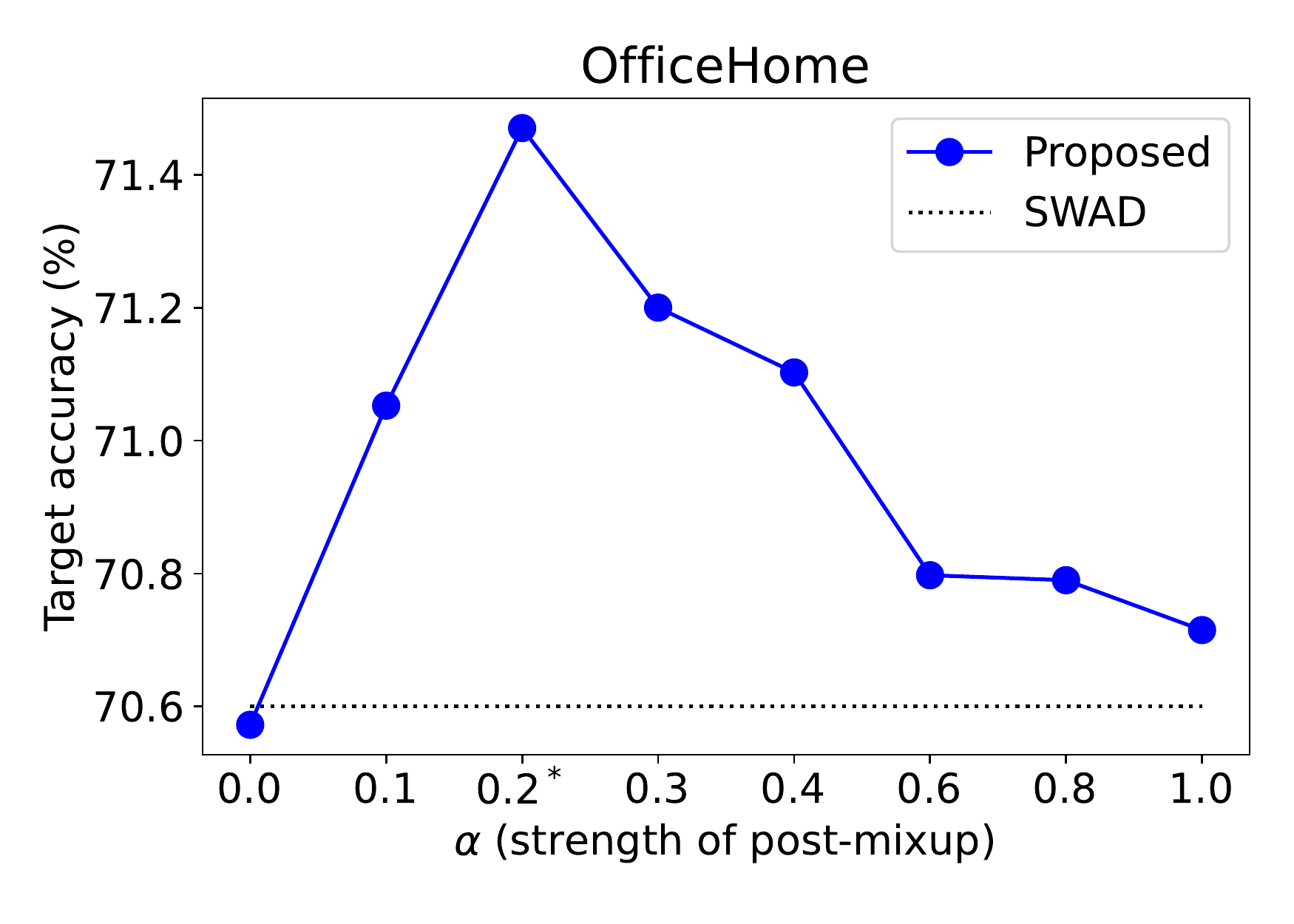}
\end{center}
\vspace{-2.0em}
\caption{Sensitivity to $\alpha$ (post-mixup strength) on PACS and OfficeHome. 
}
\label{fig:sensitivity_alpha}
\end{figure}

\noindent\textbf{Sensitivity to $\alpha$ (post-synthesis mixup strength).}\quad 
We mix up the generated amplitude images and the source domain images as in (\ref{eq:post_mixup}) to make the adversarial target domain classification task solvable.  The task becomes easier for small $\alpha$ (less impact of the generated amplitudes), and vice versa. Note that $\alpha=0$ ignores generated amplitude images completely in post-mixup, and the training becomes close to ERM learning where the only difference is that we utilise more basic augmentation (e.g., flip, rotation, color jittering). 
As shown in Fig.~\ref{fig:sensitivity_alpha}, the target test performance is not very sensitive around the best selected hyperparameters. 
See also ablation study results on the impact of post-mixup below. 

\noindent\textbf{Impact of SMCD (vs.~unsupervised MCD).}\quad
We verify the positive effect of the proposed {\em supervised MCD} loss (SMCD in (\ref{eq:modified_max_marg_obj})) that exploits the induced target domain class labels,  compared to the conventional (unsupervised) MCD loss (\ref{eq:max_marg_obj}) without using the target class labels. 
The result in Table~\ref{tab:abl_study} 
supports our claim that exploiting target class labels induced from the phase information is quite effective, improving the target generalisation performance.


\begin{table}[t!]
\vspace{-1.0em}
\caption{Ablation study: 
1) unsupervised MCD (instead of SMCD), 2) without post-mixup, 3) without SWAD, and 4) pixel-based target image generation (instead of amplitude generation). 
}
\centering
\begin{scriptsize}
\centering
\begin{tabular}{lccccc}
\toprule
 & Art & Cartoon & Photo & Sketch & Average
\\ \hline
Unsupervised MCD\Tstrut & $88.94 \pm 0.23$ & $83.83 \pm 0.19$ & $97.27 \pm 0.10$ & $81.77 \pm 0.36$ & $87.95$ \\ 
Without post-mixup\Tstrut & $88.90 \pm 0.16$ & $81.80 \pm 0.17$ & $97.43 \pm 0.14$ & $80.86 \pm 0.31$ & $87.25$ \\ 
Without SWAD\Tstrut & $84.20 \pm 0.68$ & $81.56 \pm 0.55$ & $94.83 \pm 0.12$ & $79.28 \pm 0.94$ & $84.97$ \\ 
Pixel-based generation\Tstrut & $88.85 \pm 0.15$ & $83.62 \pm 0.26$ & $97.23 \pm 0.15$ & $82.10 \pm 0.63$ & $87.95$ \\ 
\midrule
(Proposed) \ourModel{}\ \ \Tstrut & $\ {\bf 89.80 \pm 0.34} \ $ & $\ {\bf 85.16 \pm 0.65} \ $ & $\ {\bf 97.59 \pm 0.27} \ $ & $\ {\bf 84.67 \pm 0.82} \ $ & $\ {\bf 89.30} \ $ \\
\bottomrule
\end{tabular}
\end{scriptsize}
\label{tab:abl_study}
\vspace{-1.5em}
\end{table}

\noindent\textbf{Impact of post-synthesis mixup.}\quad 
We argued that our post-synthesis mixup of the generated amplitude images 
makes the class prediction task easier for the generated target domain, for the solvability of the DG problem. To verify this, we compare two models, with and without the post-mixup strategy in Table~\ref{tab:abl_study}. 
The model trained with post-mixup performs better.

\noindent\textbf{Impact of SWAD.}\quad
We adopted the SWAD model averaging scheme~\citep{swad} for improving generalisation performance. We verify the impact of the SWAD as in Table~\ref{tab:abl_study} 
where the model without SWAD has lower target test accuracy signifying the importance of the SWAD model averaging.

\noindent\textbf{Impact of amplitude generation.}\quad
The amplitude image generation in our adversarial MCD learning allows us to separate the phase and amplitude images and exploit the class labels induced by the phase information. However, one may be curious about how the model would work if we instead generate full images without phase/amplitude separation in an adversarial way. That is, we adopt a pixel-based adversarial image generator, and in turn replace our SMCD by the conventional MCD loss (since there are no class labels inducible in this strategy). We consider two generator architectures: linear (from $100$-dim input noise to full image pixels) and nonlinear (a fully connected network with one hidden layer of $100$ units), where the former slightly performs better. 
Table~\ref{tab:abl_study} 
shows that this pixel-based target image generation underperforms our amplitude generation. 

\section{Conclusion}
\vspace{-0.5em}
We tried to address the domain generalisation problem from the perspective of maximum classifier discrepancy: Improving robustness by synthesising a worst-case target domain for learning, and training the model to be robust to that domain with the (S)MCD objective. To provide an approximation to style and content separation for synthesis, the worst-case domain is synthesised in Fourier amplitude space. Our results provide a clear improvement on the state-of-the-arts on the challenging DomainBed benchmark suite.



\bibliography{main}

\begin{thebibliography}{52}
\providecommand{\natexlab}[1]{#1}
\providecommand{\url}[1]{\texttt{#1}}
\expandafter\ifx\csname urlstyle\endcsname\relax
  \providecommand{\doi}[1]{doi: #1}\else
  \providecommand{\doi}{doi: \begingroup \urlstyle{rm}\Url}\fi

\bibitem[Arjovsky et~al.(2019)Arjovsky, Bottou, Gulrajani, and Lopez-Paz]{irm}
Martin Arjovsky, L\'{e}on Bottou, Ishaan Gulrajani, and David Lopez-Paz.
\newblock {Invariant risk minimization}.
\newblock \emph{arXiv preprint arXiv:1907.02893}, 2019.

\bibitem[Balaji et~al.(2018)Balaji, Sankaranarayanan, and Chellappa]{metareg}
Yogesh Balaji, Swami Sankaranarayanan, and Rama Chellappa.
\newblock {Metareg: Towards domain generalization using meta-regularization}.
\newblock \emph{In Advances in Neural Information Processing Systems}, 2018.

\bibitem[Beery et~al.(2018)Beery, Horn, and Perona]{ti}
Sara Beery, Grant~Van Horn, and Pietro Perona.
\newblock Recognition in terra incognita.
\newblock \emph{European Conference on Computer Vision}, 2018.

\bibitem[Ben-David et~al.(2010)Ben-David, Blitzer, Crammer, Kulesza, Pereira,
  and Vaughan]{ben-david-2010}
S.~Ben-David, J.~Blitzer, K.~Crammer, A.~Kulesza, F.~Pereira, and J.~W.
  Vaughan.
\newblock A theory of learning from different domains.
\newblock \emph{Machine Learning}, 79\penalty0 (1--2):\penalty0 151--175, 2010.

\bibitem[Ben-David et~al.(2007)Ben-David, Blitzer, Crammer, and
  Pereira]{ben-david-2007}
Shai Ben-David, John Blitzer, Koby Crammer, and Fernando Pereira.
\newblock Analysis of representations for domain adaptation.
\newblock \emph{In Advances in Neural Information Processing Systems}, 2007.

\bibitem[Blanchard et~al.(2021)Blanchard, Deshmukh, Dogan, Lee, and Scott]{mtl}
Gilles Blanchard, Aniket~Anand Deshmukh, Urun Dogan, Gyemin Lee, and Clayton
  Scott.
\newblock Domain generalization by marginal transfer learning.
\newblock \emph{Journal of Machine Learning Research}, 22\penalty0
  (2):\penalty0 1--55, 2021.

\bibitem[Boyd \& Vandenberghe(2004)Boyd and Vandenberghe]{cvx}
S.~Boyd and L.~Vandenberghe.
\newblock \emph{Convex Optimization}.
\newblock Cambridge: Cambridge University Press, 2004.

\bibitem[Cha et~al.(2021)Cha, Chun, Lee, Cho, Park, Lee, and Park]{swad}
Junbum Cha, Sanghyuk Chun, Kyungjae Lee, Han-Cheol Cho, Seunghyun Park, Yunsung
  Lee, and Sungrae Park.
\newblock {SWAD: Domain Generalization by Seeking Flat Minima}.
\newblock In \emph{Advances in Neural Information Processing Systems
  (NeurIPS)}, 2021.

\bibitem[Chattopadhyay et~al.(2020)Chattopadhyay, Balaji, and Hoffman]{dmg}
Prithvijit Chattopadhyay, Yogesh Balaji, and Judy Hoffman.
\newblock {Learning to balance specificity and invariance for in and out of
  domain generalization}.
\newblock \emph{European Conference on Computer Vision}, 2020.

\bibitem[Deng et~al.(2009)Deng, Dong, Socher, Li, Li, and
  Fei-Fei]{deng2009imagenet}
Jia Deng, Wei Dong, Richard Socher, Li-Jia Li, Kai Li, and Li~Fei-Fei.
\newblock {ImageNet: A large-scale hierarchical image database}.
\newblock In \emph{IEEE Conference on Computer Vision and Pattern Recognition},
  2009.

\bibitem[Dou et~al.(2019)Dou, Castro, Kamnitsas, and Glocker]{masf}
Qi~Dou, Daniel~C Castro, Konstantinos Kamnitsas, and Ben Glocker.
\newblock {Domain generalization via model-agnostic learning of semantic
  features}.
\newblock \emph{In Advances in Neural Information Processing Systems}, 2019.

\bibitem[Fang et~al.(2013)Fang, Xu, and Rockmore]{vlcs}
Chen Fang, Ye~Xu, and Daniel~N. Rockmore.
\newblock {Unbiased metric learning: On the utilization of multiple datasets
  and web images for softening bias}.
\newblock \emph{International Conference on Computer Vision}, 2013.

\bibitem[Ganin et~al.(2016)Ganin, Ustinova, Ajakan, Germain, Larochelle,
  Laviolette, Marchand, and Lempitsky]{dann}
Yaroslav Ganin, Evgeniya Ustinova, Hana Ajakan, Pascal Germain, Hugo
  Larochelle, Fran\c{c}ois Laviolette, Mario Marchand, and Victor Lempitsky.
\newblock Domain-adversarial training of neural networks.
\newblock \emph{Journal of Machine Learning Research}, 17\penalty0
  (1):\penalty0 1--35, 2016.

\bibitem[Geirhos et~al.(2020)Geirhos, Jacobsen, Michaelis, Zemel, Brendel,
  Bethge, and Wichmann]{geirhos2020shortcut}
Robert Geirhos, J{\"o}rn-Henrik Jacobsen, Claudio Michaelis, Richard Zemel,
  Wieland Brendel, Matthias Bethge, and Felix~A Wichmann.
\newblock Shortcut learning in deep neural networks.
\newblock \emph{Nature Machine Intelligence}, 2\penalty0 (11):\penalty0
  665--673, 2020.

\bibitem[Goodfellow et~al.(2014)Goodfellow, Pouget-Abadie, Mirza, Xu,
  Warde-Farley, Ozair, Courville, and Bengio]{gan14}
I.~Goodfellow, J.~Pouget-Abadie, M.~Mirza, B.~Xu, D.~Warde-Farley, S.~Ozair,
  A.~Courville, and Y.~Bengio.
\newblock Generative adversarial nets.
\newblock \emph{In Advances in Neural Information Processing Systems}, 2014.

\bibitem[Gulrajani \& Lopez-Paz(2021)Gulrajani and Lopez-Paz]{domainbed}
Ishaan Gulrajani and David Lopez-Paz.
\newblock In search of lost domain generalization.
\newblock \emph{International Conference on Learning Representations}, 2021.

\bibitem[He et~al.(2016)He, Zhang, Ren, and Sun]{resnet}
K.~He, X.~Zhang, S.~Ren, and J.~Sun.
\newblock {Deep residual learning for image recognition}.
\newblock In \emph{IEEE/CVF Conference on Computer Vision and Pattern
  Recognition}, 2016.

\bibitem[Huang et~al.(2020)Huang, Wang, Xing, and Huang]{rsc}
Zeyi Huang, Haohan Wang, Eric~P. Xing, and Dong Huang.
\newblock Self-challenging improves cross domain generalization.
\newblock \emph{European Conference on Computer Vision}, 2020.

\bibitem[Izmailov et~al.(2018)Izmailov, Podoprikhin, Garipov, Vetrov, and
  Wilson]{swa}
Pavel Izmailov, Dmitrii Podoprikhin, Timur Garipov, Dmitry Vetrov, and
  Andrew~Gordon Wilson.
\newblock {Averaging weights leads to wider optima and better generalization}.
\newblock In \emph{Uncertainty in Artificial Intelligence}, 2018.

\bibitem[Kim et~al.(2019)Kim, Sahu, Gholami, and Pavlovic]{gpda}
Minyoung Kim, Pritish Sahu, Behnam Gholami, and Vladimir Pavlovic.
\newblock {Unsupervised Visual Domain Adaptation: A Deep Max-Margin Gaussian
  Process Approach}.
\newblock \emph{Computer Vision and Pattern Recognition}, 2019.

\bibitem[Kingma \& Ba(2015)Kingma and Ba]{adam}
Diederik~P. Kingma and Jimmy Ba.
\newblock {Adam: A method for stochastic optimization}.
\newblock \emph{International Conference on Learning Representations}, 2015.

\bibitem[Koh et~al.(2021)Koh, Sagawa, Marklund, Xie, Zhang, Balsubramani, Hu,
  Yasunaga, Phillips, Gao, Lee, David, Stavness, Guo, Earnshaw, Haque, Beery,
  Leskovec, Kundaje, Pierson, Levine, Finn, and Liang]{koh2021wilds}
Pang~Wei Koh, Shiori Sagawa, Henrik Marklund, Sang~Michael Xie, Marvin Zhang,
  Akshay Balsubramani, Weihua Hu, Michihiro Yasunaga, Richard~Lanas Phillips,
  Irena Gao, Tony Lee, Etienne David, Ian Stavness, Wei Guo, Berton Earnshaw,
  Imran Haque, Sara~M Beery, Jure Leskovec, Anshul Kundaje, Emma Pierson,
  Sergey Levine, Chelsea Finn, and Percy Liang.
\newblock Wilds: A benchmark of in-the-wild distribution shifts.
\newblock In \emph{ICML}, volume 139 of \emph{Proceedings of Machine Learning
  Research}, pp.\  5637--5664. PMLR, 2021.

\bibitem[Krueger et~al.(2020)Krueger, Caballero, Jacobsen, Zhang, Binas, Priol,
  and Courville]{vrex}
David Krueger, Ethan Caballero, Joern-Henrik Jacobsen, Amy Zhang, Jonathan
  Binas, Remi~Le Priol, and Aaron Courville.
\newblock {Out-of-distribution generalization via risk extrapolation (REx)}.
\newblock \emph{arXiv preprint arXiv:2003.00688}, 2020.

\bibitem[Li et~al.(2017)Li, Yang, Song, and Hospedales]{pacs}
Da~Li, Yongxin Yang, Yi-Zhe Song, and Timothy~M. Hospedales.
\newblock Deeper, broader and artier domain generalization.
\newblock \emph{International Conference on Computer Vision}, 2017.

\bibitem[Li et~al.(2018{\natexlab{a}})Li, Yang, Song, and Hospedales]{mldg}
Da~Li, Yongxin Yang, Yi-Zhe Song, and Timothy Hospedales.
\newblock {Learning to generalize: Meta learning for domain generalization}.
\newblock \emph{AAAI Conference on Artificial Intelligence},
  2018{\natexlab{a}}.

\bibitem[Li et~al.(2019)Li, Zhang, Yang, Liu, Song, and
  Hospedales]{li2019episodic}
Da~Li, Jianshu Zhang, Yongxin Yang, Cong Liu, Yi{-}Zhe Song, and Timothy~M.
  Hospedales.
\newblock Episodic training for domain generalization.
\newblock In \emph{ICCV}, 2019.

\bibitem[Li et~al.(2018{\natexlab{b}})Li, Pan, Wang, and Kot]{mmd}
Haoliang Li, Sinno~Jialin Pan, Shiqi Wang, and Alex~C Kot.
\newblock {Domain generalization with adversarial feature learning}.
\newblock \emph{IEEE Conference on Computer Vision and Pattern Recognition},
  2018{\natexlab{b}}.

\bibitem[Li et~al.(2018{\natexlab{c}})Li, Gong, Tian, Liu, and Tao]{cdann}
Ya~Li, Mingming Gong, Xinmei Tian, Tongliang Liu, and Dacheng Tao.
\newblock {Domain generalization via conditional invariant representations}.
\newblock \emph{AAAI Conference on Artificial Intelligence},
  2018{\natexlab{c}}.

\bibitem[Lu et~al.(2020)Lu, Yang, Zhu, Liu, Song, and Xiang]{lu2020stochastic}
Zhihe Lu, Yongxin Yang, Xiatian Zhu, Cong Liu, Yi-Zhe Song, and Tao Xiang.
\newblock Stochastic classifiers for unsupervised domain adaptation.
\newblock In \emph{Proceedings of the IEEE/CVF Conference on Computer Vision
  and Pattern Recognition}, pp.\  9111--9120, 2020.

\bibitem[Nam et~al.(2021)Nam, Lee, Park, Yoon, and Yoo]{sagnet}
Hyeonseob Nam, HyunJae Lee, Jongchan Park, Wonjun Yoon, and Donggeun Yoo.
\newblock Reducing domain gap by reducing style bias.
\newblock \emph{IEEE/CVF Conference on Computer Vision and Pattern
  Recognition}, 2021.

\bibitem[Nuriel et~al.(2021)Nuriel, Benaim, and Wolf]{padain}
Oren Nuriel, Sagie Benaim, and Lior Wolf.
\newblock {Permuted AdaIN: Reducing the bias towards global statistics in image
  classification}.
\newblock \emph{IEEE/CVF Conference on Computer Vision and Pattern
  Recognition}, 2021.

\bibitem[Oppenheim \& Lim(1981)Oppenheim and Lim]{oppenheim1981importance}
Alan~V Oppenheim and Jae~S Lim.
\newblock The importance of phase in signals.
\newblock \emph{Proceedings of the IEEE}, 69\penalty0 (5):\penalty0 529--541,
  1981.

\bibitem[Peng et~al.(2019)Peng, Bai, Xia, Huang, Saenko, and Wang]{domain_net}
Xingchao Peng, Qinxun Bai, Xide Xia, Zijun Huang, Kate Saenko, and Bo~Wang.
\newblock Moment matching for multi-source domain adaptation.
\newblock \emph{International Conference on Computer Vision}, 2019.

\bibitem[Sagawa et~al.(2020)Sagawa, Koh, Hashimoto, and Liang]{group_dro}
Shiori Sagawa, Pang~Wei Koh, Tatsunori~B. Hashimoto, and Percy Liang.
\newblock {Distributionally robust neural networks}.
\newblock \emph{International Conference on Learning Representations}, 2020.

\bibitem[Saito et~al.(2018)Saito, Watanabe, Ushiku, and Harada]{saito2018}
Kuniaki Saito, Kohei Watanabe, Yoshitaka Ushiku, and Tatsuya Harada.
\newblock Maximum classifier discrepancy for unsupervised domain adaptation.
\newblock \emph{Computer Vision and Pattern Recognition}, 2018.

\bibitem[Seo et~al.(2020)Seo, Suh, Kim, Han, and Han]{dson}
Seonguk Seo, Yumin Suh, Dongwan Kim, Jongwoo Han, and Bohyung Han.
\newblock {Learning to optimize domain specific normalization for domain
  generalization}.
\newblock \emph{European Conference on Computer Vision}, 2020.

\bibitem[Shankar et~al.(2018)Shankar, Piratla, Chakrabarti, Chaudhuri, Jyothi,
  and Sarawagi]{shankar2018crossgrad}
Shiv Shankar, Vihari Piratla, Soumen Chakrabarti, Siddhartha Chaudhuri, Preethi
  Jyothi, and Sunita Sarawagi.
\newblock Generalizing across domains via cross-gradient training.
\newblock In \emph{ICLR}, 2018.

\bibitem[Sun \& Saenko(2016)Sun and Saenko]{coral}
Baochen Sun and Kate Saenko.
\newblock {Deep coral: Correlation alignment for deep domain adaptation}.
\newblock \emph{European Conference on Computer Vision}, 2016.

\bibitem[Vedantam et~al.(2021)Vedantam, Lopez-Paz, and Schwab]{investigate_dg}
Ramakrishna Vedantam, David Lopez-Paz, and David~J. Schwab.
\newblock {An Empirical Investigation of Domain Generalization with Empirical
  Risk Minimizers}.
\newblock In \emph{Advances in Neural Information Processing Systems
  (NeurIPS)}, 2021.

\bibitem[Venkateswara et~al.(2017)Venkateswara, Eusebio, Chakraborty, and
  Panchanathan]{office_home}
Hemanth Venkateswara, Jose Eusebio, Shayok Chakraborty, and Sethuraman
  Panchanathan.
\newblock Deep hashing network for unsupervised domain adaptation.
\newblock \emph{Computer Vision and Pattern Recognition}, 2017.

\bibitem[Wang et~al.(2020{\natexlab{a}})Wang, Yu, Li, Fu, and Heng]{eisnet}
Shujun Wang, Lequan Yu, Caizi Li, Chi-Wing Fu, and Pheng-Ann Heng.
\newblock {Learning from extrinsic and intrinsic supervisions for domain
  generalization}.
\newblock \emph{European Conference on Computer Vision}, 2020{\natexlab{a}}.

\bibitem[Wang et~al.(2020{\natexlab{b}})Wang, Li, and Kot]{imixup_3}
Yufei Wang, Haoliang Li, and Alex~C Kot.
\newblock {Heterogeneous domain generalization via domain mixup}.
\newblock \emph{IEEE International Conference on Acoustics, Speech and Signal
  Processing}, 2020{\natexlab{b}}.

\bibitem[Xu et~al.(2020)Xu, Zhang, Ni, Li, Wang, Tian, and Zhang]{imixup_1}
Minghao Xu, Jian Zhang, Bingbing Ni, Teng Li, Chengjie Wang, Qi~Tian, and
  Wenjun Zhang.
\newblock {Adversarial domain adaptation with domain mixup}.
\newblock \emph{AAAI Conference on Artificial Intelligence}, 2020.

\bibitem[Xu et~al.(2021)Xu, Zhang, Zhang, Wang, and Tian]{fact}
Qinwei Xu, Ruipeng Zhang, Ya~Zhang, Yanfeng Wang, and Qi~Tian.
\newblock {A Fourier-based Framework for Domain Generalization}.
\newblock In \emph{IEEE/CVF Conference on Computer Vision and Pattern
  Recognition (CVPR)}, June 2021.

\bibitem[Yan et~al.(2020)Yan, Song, Li, Zou, and Ren]{imixup_2}
Shen Yan, Huan Song, Nanxiang Li, Lincan Zou, and Liu Ren.
\newblock {Improve unsupervised domain adaptation with mixup training}.
\newblock \emph{arXiv preprint arXiv:2001.00677}, 2020.

\bibitem[Zhang et~al.(2018)Zhang, Cisse, Dauphin, and Lopez-Paz]{mixup}
Hongyi Zhang, Moustapha Cisse, Yann~N. Dauphin, and David Lopez-Paz.
\newblock {Mixup: Beyond Empirical Risk Minimization}.
\newblock \emph{International Conference on Learning Representations}, 2018.

\bibitem[Zhang et~al.(2020)Zhang, Marklund, Gupta, Levine, and Finn]{arm}
Marvin Zhang, Henrik Marklund, Abhishek Gupta, Sergey Levine, and Chelsea Finn.
\newblock {Adaptive risk minimization: A meta-learning approach for tackling
  group shift}.
\newblock \emph{arXiv preprint arXiv:2007.02931}, 2020.

\bibitem[Zhang et~al.(2022)Zhang, Li, Li, Jia, and Zhang]{efdmix}
Yabin Zhang, Minghan Li, Ruihuang Li, Kui Jia, and Lei Zhang.
\newblock {Exact Feature Distribution Matching for Arbitrary Style Transfer and
  Domain Generalization}.
\newblock \emph{IEEE/CVF Conference on Computer Vision and Pattern
  Recognition}, 2022.

\bibitem[Zhao et~al.(2020)Zhao, Gong, Liu, Fu, and Tao]{er}
Shanshan Zhao, Mingming Gong, Tongliang Liu, Huan Fu, and Dacheng Tao.
\newblock {Domain generalization via entropy regularization}.
\newblock \emph{In Advances in Neural Information Processing Systems}, 2020.

\bibitem[Zhou et~al.(2020)Zhou, Yang, Hospedales, and Xiang]{zhou2020deepDG}
Kaiyang Zhou, Yongxin Yang, Timothy Hospedales, and Tao Xiang.
\newblock Deep domain-adversarial image generation for domain generalisation.
\newblock In \emph{Proceedings of the AAAI Conference on Artificial
  Intelligence}, 2020.

\bibitem[Zhou et~al.(2021{\natexlab{a}})Zhou, Liu, Qiao, Xiang, and
  Loy]{zhou2021domainGenSurvey}
Kaiyang Zhou, Ziwei Liu, Yu~Qiao, Tao Xiang, and Chen~Change Loy.
\newblock Domain generalization: A survey.
\newblock \emph{arXiv preprint arXiv:2103.02503}, 2021{\natexlab{a}}.

\bibitem[Zhou et~al.(2021{\natexlab{b}})Zhou, Yang, Qiao, and Xiang]{mixstyle}
Kaiyang Zhou, Yongxin Yang, Yu~Qiao, and Tao Xiang.
\newblock Domain generalization with mixstyle.
\newblock \emph{International Conference on Learning Representations},
  2021{\natexlab{b}}.

\end{thebibliography}
\bibliographystyle{iclr2023_conference}

\clearpage

\appendix
\section{Appendix}

The Appendix consists of the following contents:
\begin{itemize}
\item Implementation Details (Sec.~\ref{sec:impl_details})
\item Full Results (Sec.~\ref{sec:full_results})
\item Derivation of ELBO in Variational Inference (Sec.~\ref{sec:derivations})
\item Additional Experimental Results (Sec.~\ref{sec:extra_expmt})
%
\end{itemize}

\subsection{Implementation Details}\label{sec:impl_details}

We adopt the ResNet50~\citep{resnet} architecture (removing the final classification layer) as the feature extractor network. For the amplitude generator network, we have tested several fully-connected network architectures with different numbers of hidden layers and hidden units, and the simple linear network peformed the best. The input noise dimension for the generator is chosen as $100$. The covariance matrices of the variational parameters are restricted to be diagonal. The number of MC samples from $Q_\lambda(W)$ in the ELBO optimisation is chosen as 50. 

The optimisation hyperparameters are chosen by the same strategy as~\citep{swad}, where we employ the Adam optimiser~\citep{adam} with learning rate $5\times 10^{-5}$, and no dropout, weight decay used. The batch size was 32 (for each training domain) in ERM/SWAD~\citep{swad}, but we halved it in our model since the remaining half are constructed by the adversarial target generation. The standard basic data augmentation is also applied to the input images. Following the suggestion from~\citep{swad}, we run our model up to 5000 iterations for all datasets except for DomainNet. But the algorithm may stop earlier before the maximum iterations if SWAD termination condition is met (See Sec.~\ref{sec:swa_alg} below). Since DomainNet is a large-scale dataset, and it is required to have a even larger number of iterations to go through the entire data at least several times. In~\citep{swad}, they used 15000 iterations which roughly corresponds to 3 to 10 data epochs. In our model, since we halved the number of input images in the batch, in order to have the same training epochs as~\citep{swad}, we increase it up to 30000 iterations for DomainNet. The details of the SWAD implementation follows in the next section. 


\subsubsection{SWAD Model Averaging}\label{sec:swa_alg}

We adopt the SWAD model averaging strategy~\citep{swad} to have a more robust model that is less affected by overfitting. We apply the SWAD to the feature extractor network parameters $\theta$ and the variational parameters $\lambda$, but not the adversarial generator network. Since SWAD is an important component in our model, we provide more details here. 

SWAD is motivated from {\em stochastic weight averaging} (SWA)~\citep{swa}, however, unlike SWA's model averaging for every {\em epoch}, SWAD takes dense model averaging for every (batch) {\em iteration}. A key component of the SWAD algorithm is to determine the model averaging regime, the interval of iterations for which the model averaging is performed. This regime is aimed to avoid overfitting, and known as overfit-aware model averaging. The regime is specified by the start and end iteration numbers, $t_s$ and $t_e$, respectively, and we take model averaging for iterations $t \in [t_s,t_e]$, that is,
\begin{align}
\theta_{SWAD} = \frac{1}{t_e-t_s+1} \sum_{t=t_s}^{t_e} \theta^t, \ \ \ \ 
\lambda_{SWAD} = \frac{1}{t_e-t_s+1} \sum_{t=t_s}^{t_e} \lambda^t,
\end{align}
where $\theta^t$ and $\lambda^t$ are the model parameters after iteration $t$. Here $(\theta_{SWAD},\lambda_{SWAD})$ are the final model parameters returned by the training algorithm. 

Now we describe how the regime is determined. Ideally, we expect the intermediate models during the interval $[t_s,t_e]$ to be overfit-free, having high generalisation performance. To this end, we evaluate the model on the validation set (held out from the source domain training data), and denote the validation loss of the $t$-th model by $l_{val}^t$. Then the start iteration of the regime, $t_s$ is determined by the first $t$ where the validation loss is not improved for the next $N_s$ iterations (e.g., $N_s=3$). That is,
\begin{align}
t_s = \min\{t-N_s+1 \ \vert \ l_{val}^{t-N_s+1} \leq l_{val}^{t}, l_{val}^{t-1}, \dots, l_{val}^{t-N_s+1} \}.
\label{eq:t_s}
\end{align}
Once we find $t_s$, we compute the (average) starting validation loss,
\begin{align}
\overline{l}_{val} = \frac{\sum_{t=t_s}^{t_s+N_s-1} l_{val}^t}{N_s},
\label{eq:l_avg}
\end{align}
which is used as a reference when we decide the end iteration $t_e$. As we enter the regime, we start model averaging every iteration. To determine when to stop, we inspect the validation losses to see if the model starts overfitting. Specifically, if the validation losses are consecutively greater than $\overline{l}_{val}$ by some margin, we regard it as overfit signal. That is,
\begin{align}
t_e = \min\{t - N_e \ \vert \ l_{val}^t, l_{val}^{t-1}, \dots, l_{val}^{t-N_e+1} > r \cdot \overline{l}_{val} \},
\label{eq:t_e}
\end{align}
where $r$ and $N_e$ are user-driven hyperparameters (e.g., $r=1.3$, $N_e=6$).

The pseudo code of our AGFA algorithm with the SWAD strategy is summarised in Alg.~\ref{alg:main_swad}. 
There are three hyperparameters in SWAD, ($N_s$, $N_e$, $r$), and following~\citep{swad}, we use $N_s=3$, $N_e=6$, $r=1.3$ for all datasets in DomainBed except $r=1.2$ for VLCS. 
One technical issue is that evaluating the validation loss every iteration is computationally demanding. Similarly as~\citep{swad}, we compute the validation loss at every $V$-th iterations (e.g., $V=50$ for VLCS, $V=500$ for DomainNet, and $V=100$ for the rest) although the model averaging is still performed every iteration. Accordingly, the equations (\ref{eq:t_s}), (\ref{eq:l_avg}), and (\ref{eq:t_e}) need to be changed where essentially all iteration numbers in those equations should be changed to multiples of $V$. The model averaging in Alg.~\ref{alg:main_swad} is implemented by the running (online) average and the use of (FIFO) queue data structures similarly as~\citep{swad}, which does not incur significant extra computational overhead. 

\newcommand\inlineeqno{\stepcounter{equation}\ (\theequation)}
\newcommand{\INDSTATE}[1][1]{\STATE\hspace{#1\algorithmicindent}}
\begin{algorithm}[t!]
\caption{AGFA Algorithm with SWAD Model Averaging.
}
\label{alg:main_swad}
\begin{algorithmic}
\STATE \textbf{Input:} Source data $S$, SMCD trade-off $\eta$, post-mixup $\alpha$, and learning rate $\gamma$, and \\
\ \ \ \ \ \ \ \ \ \ \ SWAD hyperparameters $N_s$, $N_e$, $r$.
\STATE \textbf{Initialise:} $\theta$ (feature extractor), $\lambda$ (variational parameters), and $\nu$ (generator). \\
\ \ \ \ \ \ \ \ \ \ \ \ \ \ \ (flag) SWAD-Regime-Entered $\leftarrow FALSE$, (iteration) $t\leftarrow 0$.
\STATE \textbf{Repeat:}
    \INDSTATE[1] 0. Sample a minibatch $S_B = \{(x^S_i,y^S_i)\}_{i=1}^n$ from $S$.
    \INDSTATE[1] 1. Prepare  $\{(A^S_i,P^S_i)\}_{i=1}^n$ by Fourier transform $A^S_i \angle P^S_i = \mathcal{F}(x^S_i)$.
    \INDSTATE[1] 2. Generate amplitude images $A^G_i = G_\nu(\epsilon_i)$,  $\epsilon_i\sim\mathcal{N}(0,I)$ for $i=1,\dots,n$.
    \INDSTATE[1] 3. Post-mixup: $A^G_i \leftarrow  \lambda A^G_i + (1-\lambda) A^S_i$,  $\lambda\sim\textrm{Uniform}(0,\alpha)$.
    \INDSTATE[1] 4. Construct a target batch $T_B=\{(x^T_i,y^T_i)\}_{i=1}^n$:  $x^T_i=\mathcal{F}^{-1}(A^G_i \angle P^S_i)$,  $y^T_i=y^S_i$.
    \INDSTATE[1] 5. Evaluate $\mathcal{L}_{model} := -\textrm{ELBO}(\lambda,\theta;S_B) + \eta \textrm{SMCD}(\theta;T_B)$.
    \INDSTATE[1] 6. Update the model and variational parameters: $(\lambda,\theta) \leftarrow (\lambda,\theta) - \gamma \nabla_{(\lambda,\theta)}\mathcal{L}_{model}$.
    \INDSTATE[1] 7. Evaluate $\mathcal{L}_{gen} := -\textrm{SMCD}(\theta;T_B)$.
    \INDSTATE[1] 8. Update the generator network: $\nu \leftarrow \nu - \gamma \nabla_\nu \mathcal{L}_{gen}$. 
    \INDSTATE[1] 9. (SWAD procedure) \\
\ \ \ \ \ \ \ \ \ \ \ $t \leftarrow t + 1$, $(\lambda^t,\theta^t) \leftarrow (\lambda,\theta)$.  \\
\ \ \ \ \ \ \ \ \ \ \ If SWAD-Regime-Entered $== FALSE$: \\
\ \ \ \ \ \ \ \ \ \ \ \ \ \ \ \ \ If $l_{val}^{t-N_s+1} = \min_{0\leq t' < N_s} l_{val}^{t-t'}$:\\
\ \ \ \ \ \ \ \ \ \ \ \ \ \ \ \ \ \ \ \ \ \ \ $t_s \leftarrow t-N_s+1$, $\overline{l}_{val} \leftarrow \frac{1}{N_s}\sum_{t'=0}^{N_s-1} l_{val}^{t-t'}$. \\
\ \ \ \ \ \ \ \ \ \ \ \ \ \ \ \ \ \ \ \ \ \ \ SWAD-Regime-Entered $\leftarrow True$.\\
\ \ \ \ \ \ \ \ \ \ \ Else: \\
\ \ \ \ \ \ \ \ \ \ \ \ \ \ \ \ \ If $r \cdot \overline{l}_{val} < \min_{0\leq t' < N_e} l_{val}^{t-t'}$:\\
\ \ \ \ \ \ \ \ \ \ \ \ \ \ \ \ \ \ \ \ \ \ \ $t_e \leftarrow t-N_e$. \\
\ \ \ \ \ \ \ \ \ \ \ \ \ \ \ \ \ \ \ \ \ \ \ Return $\theta_{SWAD} = \frac{1}{t_e-t_s+1} \sum_{t=t_s}^{t_e} \theta^t$, $\lambda_{SWAD} = \frac{1}{t_e-t_s+1} \sum_{t=t_s}^{t_e} \lambda^t$.
\end{algorithmic}
\end{algorithm}

\subsection{Full Results}\label{sec:full_results}

The full results (test errors on individual target domains) on DomainBed datasets are summarised in Table~\ref{tab:pacs} (PACS), Table~\ref{tab:vlcs} (VLCS), Table~\ref{tab:oh} (OfficeHome), Table~\ref{tab:ti} (TerraIncognita), and Table~\ref{tab:dn} (DomainNet).

\begin{figure}
\vspace{-3.0em}
\begin{center}
%
\centering
\includegraphics[trim = 2mm 2mm 5mm 10mm, clip, scale=0.375]{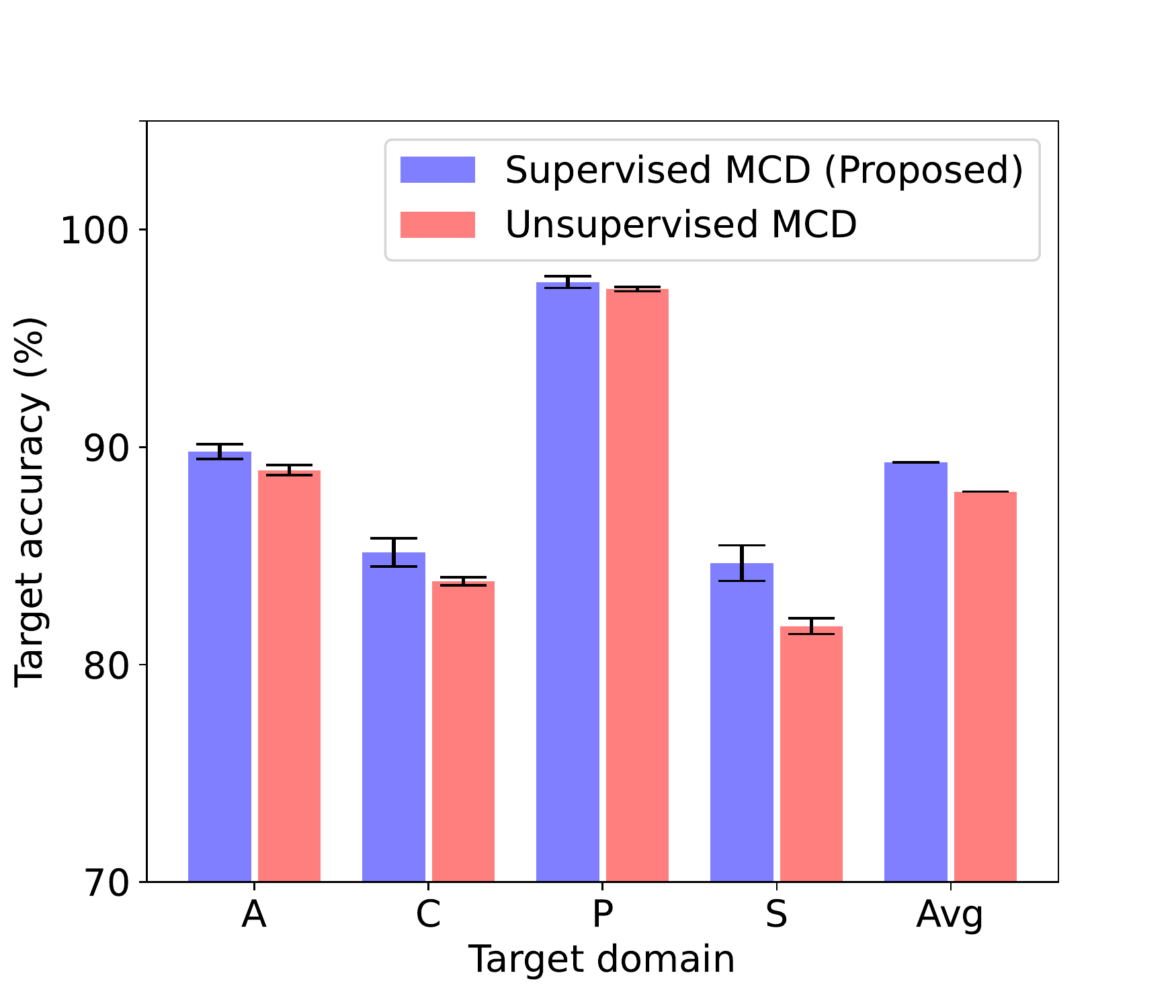}
\includegraphics[trim = 2mm 2mm 5mm 10mm, clip, scale=0.375]{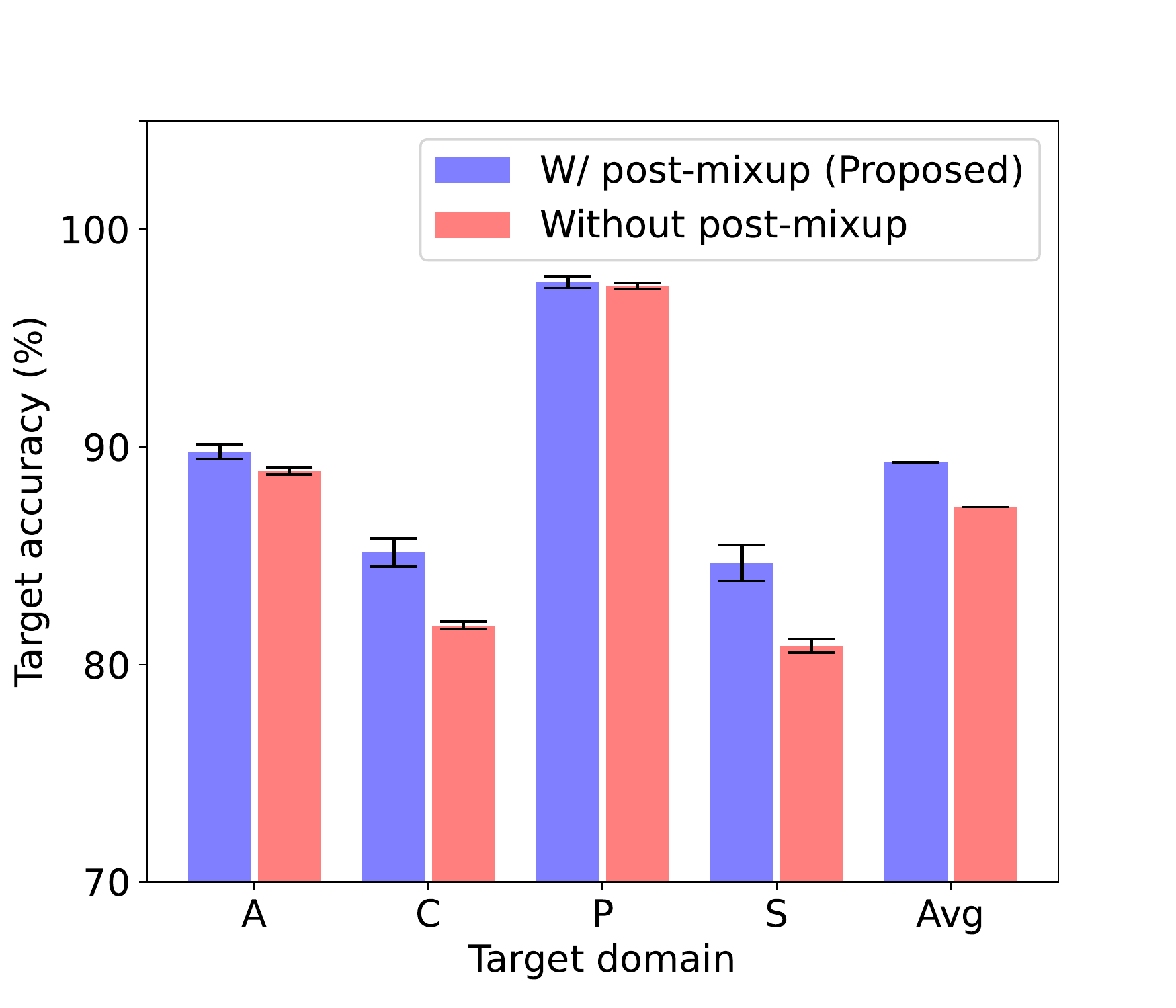} \\
\vspace{-0.3em}
\includegraphics[trim = 2mm 2mm 5mm 12mm, clip, scale=0.375]{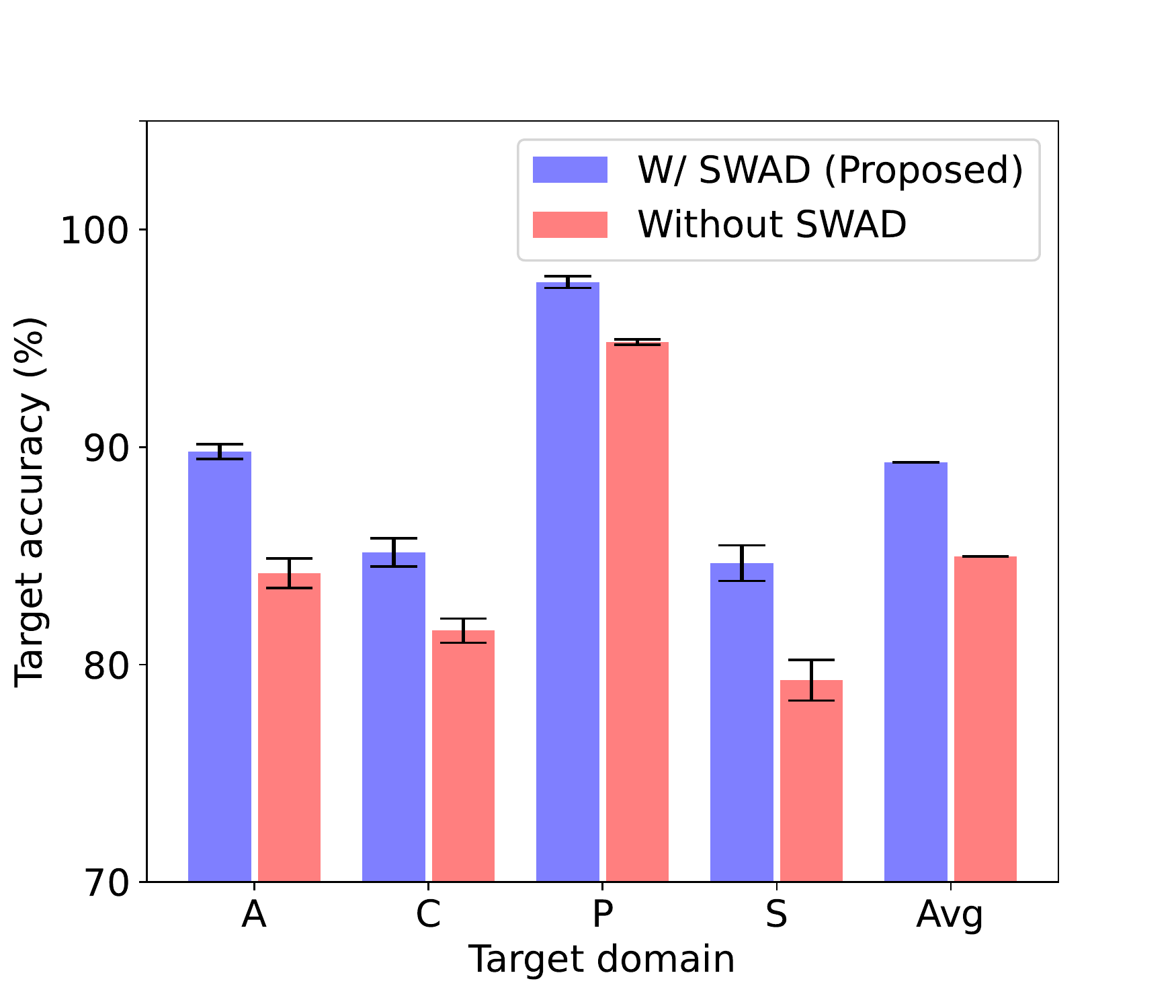}
\includegraphics[trim = 2mm 2mm 5mm 12mm, clip, scale=0.375]{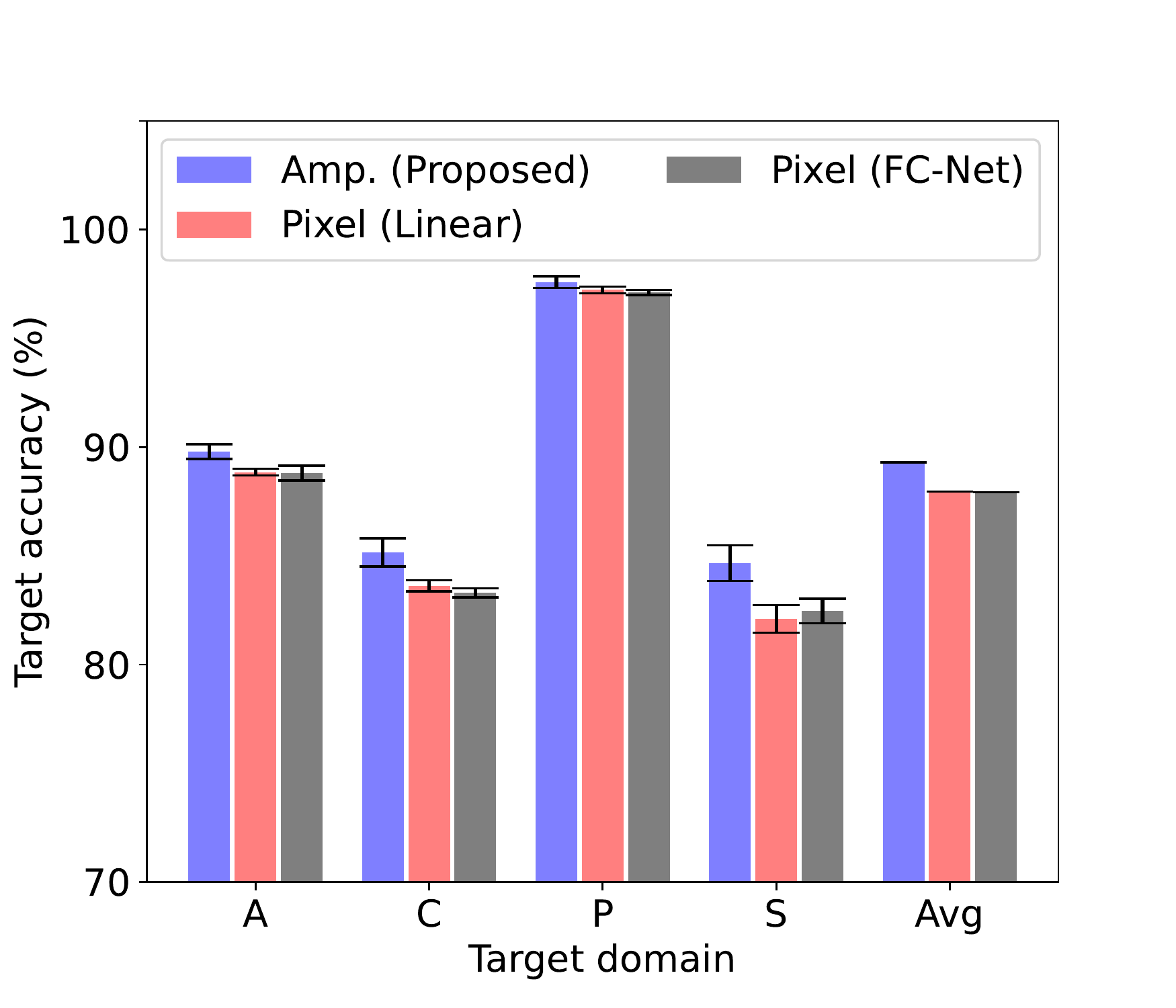}
\end{center}
\vspace{-1.0em}
\caption{Ablation study of four different modeling choices: SMCD, post-mixup, SWAD, and amplitude generation (instead of pixel-based target image generation). 
}
\label{fig:abl_study}
\end{figure}

We also show the full results of the sensitivity analysis in Table~\ref{tab:abl_eta} (the SMCD loss trade-off $\eta$) and Table~\ref{tab:abl_alpha} (the post-synthesis mixup strength $\alpha$). Moreover, we visualise in Fig.~\ref{fig:abl_study} the ablation study results for the four different modeling choices: 1) Impact of SMCD (vs.~conventional unsupervised MCD), 2) Impact of post-synthesis mixup, 3) Impact of SWAD, and 4) Impact of amplitude generation (vs.~pixel-based image generation). For the pixel-based image generation, we consider two generator architectures: linear (from $100$-dim input noise to full image pixels) and nonlinear (a fully connected network with one hidden layer of $100$ units).

\begin{figure}[t!]
\begin{center}
%
\centering
\includegraphics[trim = 2mm 2mm 6mm 4mm, clip, scale=0.375 
]{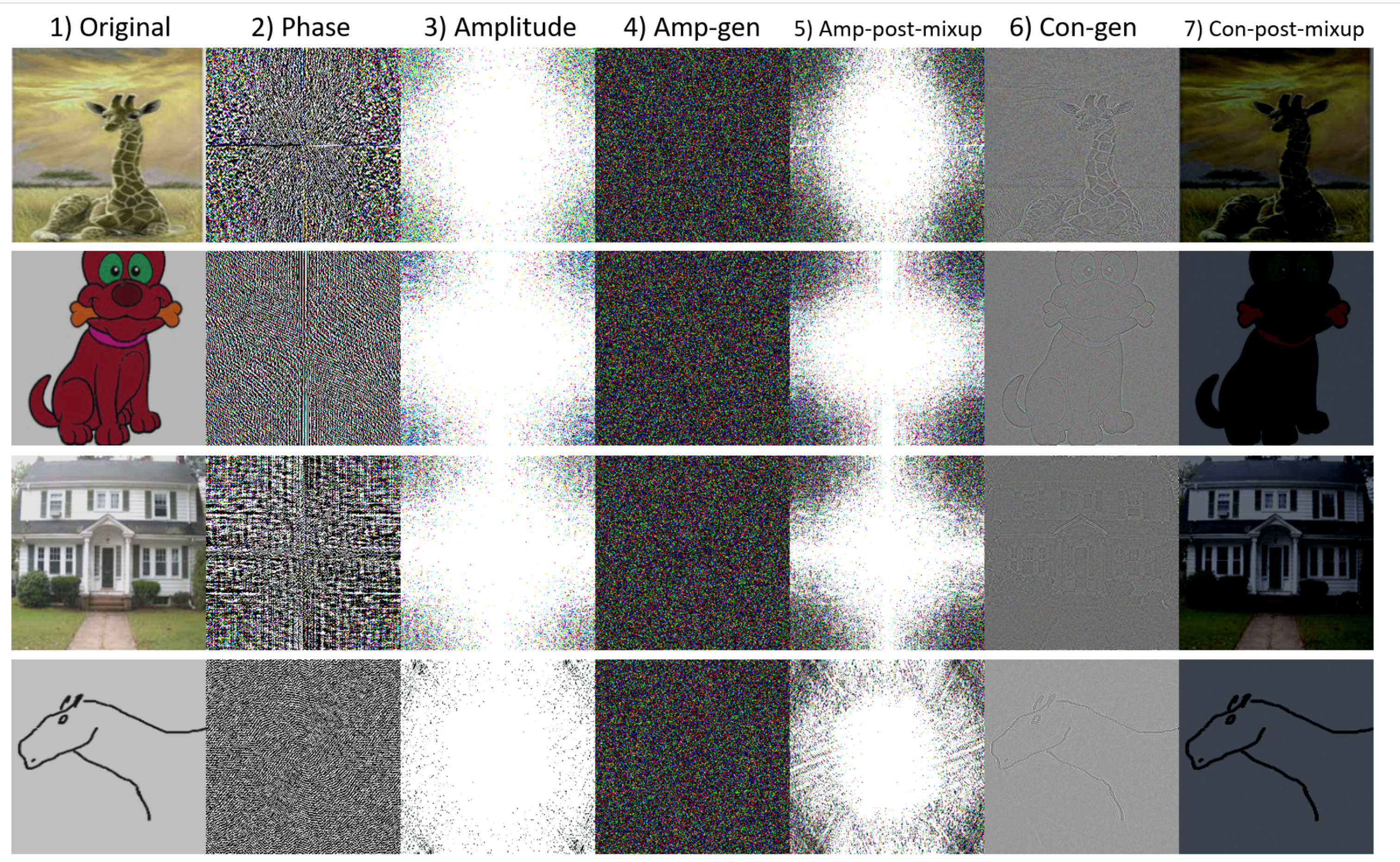}
\end{center}
\vspace{-1.5em}
\caption{Visualisation of the generated amplitude and constructed images. The columns are (from left to right): 1) original image, 2) phase and 3) amplitude spectra after Fourier transform, 4) generated amplitude image, 5) post-mixup of 3 and 4, 6) constructed image from phase in 2) and generated amplitude image in 4) (by inverse Fourier transform), and 7) constructed image from phase 2 and the post-mixup amplitude 5.
}
\label{fig:vis_acps}
\end{figure}

\noindent\textbf{Visualisation of generated adversarial images.}\quad\label{sec:vis_adv_images}
We visualise in Fig.~\ref{fig:vis_acps} some synthesised amplitude images and constructed target domain images from the learned model on the PACS dataset. Although the generated amplitude images visually look like random noise, they appear to have the effect of attenuating high frequency spectra (shown as darker pixels in the fifth column) when combined with the source domain amplitude images by post-mixup. The constructed images from the generated amplitude images alone without post-mixup (sixth column) look a lot like edge detection maps, whereas the post-mixup constructed ones (seventh column) remain visually similar to the original source domain images, promoting DG solvability.

\definecolor{lor}{rgb}{1,0.85,0}
\definecolor{or}{rgb}{1,0.60,0}
\definecolor{dor}{rgb}{1,0.20,0}

\begin{table}[t!]
\vspace{-3.0em}
\centering
\caption{Average accuracies on PACS. 
Note: $^\dagger$ indicates that the results are excerpted from the published papers or~\citep{domainbed}. Our own runs are reported without $^\dagger$. 
FACT~\citep{fact} adopted a slightly different data/domain split from DomainBed's, explaining discrepancy. 
}
\label{tab:pacs}
\begin{scriptsize}
\centering
\scalebox{0.95}{
\begin{tabular}{lccccc}
\toprule
Algorithm \ \ \ \ \ \ \ \ \ \ \ \ \ \ & \ \ \ \ \ \ A \ \ \ \ \ \ & \ \ \ \ \ \ C \ \ \ \ \ \ & \ \ \ \ \ \ P \ \ \ \ \ \ & \ \ \ \ \ \ S \ \ \ \ \ \ & \ \ Avg \ \ \\
\midrule
MASF~\citep{masf}$^\dagger$\Tstrut & $82.9$ & $80.5$ & $95.0$ & $72.3$ & $82.7$ \\ 
DMG~\citep{dmg}$^\dagger$\Tstrut & $82.6$ & $78.1$ & $94.5$ & $78.3$ & $83.4$ \\ 
MetaReg~\citep{metareg}$^\dagger$\Tstrut & $87.2$ & $79.2$ & $97.6$ & $70.3$ & $83.6$ \\ 
ER~\citep{er}$^\dagger$\Tstrut & $87.5$ & $79.3$ & ${\bf 98.3}$ & $76.3$ & $85.3$ \\ 
pAdaIN~\citep{padain}$^\dagger$\Tstrut & $85.8$ & $81.1$ & $97.2$ & $77.4$ & $85.4$ \\ 
EISNet~\citep{eisnet}$^\dagger$\Tstrut & $86.6$ & $81.5$ & $97.1$ & $78.1$ & $85.8$ \\ 
DSON~\citep{dson}$^\dagger$\Tstrut & $87.0$ & $80.6$ & $96.0$ & $82.9$ & $86.6$ \\ 
ERM~\citep{swad}$^\dagger$\Tstrut & $85.7 \pm 0.6$ & $77.1 \pm 0.8$ & $97.4 \pm 0.4$ & $76.6 \pm 0.7$ & $84.2$ \\ 
IRM~\citep{irm}$^\dagger$\Tstrut & $84.8 \pm 1.3$ & $76.4 \pm 1.1$ & $96.7 \pm 0.6$ & $76.1 \pm 1.0$ & $83.5$ \\ 
GroupDRO~\citep{group_dro}$^\dagger$\Tstrut & $83.5 \pm 0.9$ & $79.1 \pm 0.6$ & $96.7 \pm 0.3$ & $78.3 \pm 2.0$ & $84.4$ \\ 
I-Mixup~\citep{imixup_1,imixup_2,imixup_3}$^\dagger$\Tstrut & $86.1 \pm 0.5$ & $78.9 \pm 0.8$ & $97.6 \pm 0.1$ & $75.8 \pm 1.8$ & $84.6$ \\ 
MLDG~\citep{mldg}$^\dagger$\Tstrut & $85.5 \pm 1.4$ & $80.1 \pm 1.7$ & $97.4 \pm 0.3$ & $76.6 \pm 1.1$ & $84.9$ \\ 
CORAL~\citep{coral}$^\dagger$\Tstrut & $88.3 \pm 0.2$ & $80.0 \pm 0.5$ & $97.5 \pm 0.3$ & $78.8 \pm 1.3$ & $86.2$ \\ 
MMD~\citep{mmd}$^\dagger$\Tstrut & $86.1 \pm 1.4$ & $79.4 \pm 0.9$ & $96.6 \pm 0.2$ & $76.5 \pm 0.5$ & $84.7$ \\ 
DANN~\citep{dann}$^\dagger$\Tstrut & $86.4 \pm 0.8$ & $77.4 \pm 0.8$ & $97.3 \pm 0.4$ & $73.5 \pm 2.3$ & $83.7$ \\
CDANN~\citep{cdann}$^\dagger$\Tstrut & $84.6 \pm 1.8$ & $75.5 \pm 0.9$ & $96.8 \pm 0.3$ & $73.5 \pm 0.6$ & $82.6$ \\ 
MTL~\citep{mtl}$^\dagger$\Tstrut & $87.5 \pm 0.8$ & $77.1 \pm 0.5$ & $96.4 \pm 0.8$ & $77.3 \pm 1.8$ & $84.6$ \\ 
SagNet~\citep{sagnet}$^\dagger$\Tstrut & $87.4 \pm 1.0$ & $80.7 \pm 0.6$ & $97.1 \pm 0.1$ & $80.0 \pm 0.4$ & $86.3$ \\
ARM~\citep{arm}$^\dagger$\Tstrut & $86.8 \pm 0.6$ & $76.8 \pm 0.5$ & $97.4 \pm 0.3$ & $79.3 \pm 1.2$ & $85.1$ \\
VREx~\citep{vrex}$^\dagger$\Tstrut & $86.0 \pm 1.6$ & $79.1 \pm 0.6$ & $96.9 \pm 0.5$ & $77.7 \pm 1.7$ & $84.9$ \\
RSC~\citep{rsc}$^\dagger$\Tstrut & $85.4 \pm 0.8$ & $79.7 \pm 1.8$ & $97.6 \pm 0.3$ & $78.2 \pm 1.2$ & $85.2$ \\
Mixstyle~\citep{mixstyle}$^\dagger$\Tstrut & $86.8 \pm 0.5$ & $79.0 \pm 1.4$ & $96.6 \pm 0.1$ & $78.5 \pm 2.3$ & $85.2$ \\
FACT~\citep{fact}$^\dagger$\Tstrut & $89.6 \pm 0.5$ & $81.8 \pm 0.2$ & $96.8 \pm 0.1$ & $84.5 \pm 0.8$ & $88.2$
\\ 
FACT~\citep{fact}\Tstrut & $87.8 \pm 0.2$ & $80.5 \pm 1.1$ & $96.2 \pm 0.2$ & $81.2 \pm 0.6$ & $86.4$ \\ 
Amp-Mixup~\citep{fact}\Tstrut & $84.7 \pm 0.6$ & $81.0 \pm 1.1$ & $95.0 \pm 0.2$ & $78.1 \pm 1.0$ & $84.7$ \\ 
\hline
SWAD~\citep{swad}$^\dagger$\Tstrut & $89.3 \pm 0.2$ & $83.4 \pm 0.6$ & $97.3 \pm 0.3$ & $82.5 \pm 0.5$ & $88.1$ \\ 
FACT$+$SWAD\Tstrut & $89.6 \pm 0.8$ & $82.5 \pm 0.3$ & $96.6 \pm 0.2$ & $83.8 \pm 0.8$ & $88.1$ \\ 
Amp-Mixup$+$SWAD\Tstrut & $88.7 \pm 0.1$ & $83.2 \pm 0.4$ & $96.4 \pm 0.1$ & $84.1 \pm 0.5$ & $88.1$ \\ 
\hline
(Proposed) \ourModel{}\Tstrut & ${\bf 89.8 \pm 0.3}$ & ${\bf 85.2 \pm 0.6}$ & $97.6 \pm 0.3$ & ${\bf 84.7 \pm 0.8}$ & ${\bf 89.3}$ \\
\bottomrule
\end{tabular}
}
\end{scriptsize}
\end{table}

\begin{table}[t!]
\centering
\caption{Average accuracies on VLCS. The same interpretation as Table~\ref{tab:pacs}.
}
\label{tab:vlcs}
\begin{scriptsize}
\centering
\scalebox{0.95}{
\begin{tabular}{lccccc}
\toprule
Algorithm \ \ \ \ \ \ \ \ \ \ & \ \ \ C \ \ \ & \ \ \ L \ \ \ & \ \ \ S \ \ \ & \ \ \ V \ \ \ & \ \ Avg \ \ \\
\midrule
%
ERM~\citep{swad}$^\dagger$\Tstrut & $98.0 \pm 0.3$ & $64.7 \pm 1.2$ & $71.4 \pm 1.2$ & $75.2 \pm 1.6$ & $77.3$ \\ 
IRM~\citep{irm}$^\dagger$\Tstrut & $98.6 \pm 0.1$ & $64.9 \pm 0.9$ & $73.4 \pm 0.6$ & $77.3 \pm 0.9$ & $78.6$ \\ 
GroupDRO~\citep{group_dro}$^\dagger$\Tstrut & $97.3 \pm 0.3$ & $63.4 \pm 0.9$ & $69.5 \pm 0.8$ & $76.7 \pm 0.7$ & $76.7$ \\ 
I-Mixup~\citep{imixup_1,imixup_2,imixup_3}$^\dagger$\Tstrut & $98.3 \pm 0.6$ & $64.8 \pm 1.0$ & $72.1 \pm 0.5$ & $74.3 \pm 0.8$ & $77.4$ \\ 
MLDG~\citep{mldg}$^\dagger$\Tstrut & $97.4 \pm 0.2$ & $65.2 \pm 0.7$ & $71.0 \pm 1.4$ & $75.3 \pm 1.0$ & $77.2$ \\ 
CORAL~\citep{coral}$^\dagger$\Tstrut & $98.3 \pm 0.1$ & ${\bf 66.1 \pm 1.2}$ & $73.4 \pm 0.3$ & $77.5 \pm 1.2$ & $78.8$ \\ 
MMD~\citep{mmd}$^\dagger$\Tstrut & $97.7 \pm 0.1$ & $64.0 \pm 1.1$ & $72.8 \pm 0.2$ & $75.3 \pm 3.3$ & $77.5$ \\ 
DANN~\citep{dann}$^\dagger$\Tstrut & ${\bf 99.0 \pm 0.3}$ & $65.1 \pm 1.4$ & $73.1 \pm 0.3$ & $77.2 \pm 0.6$ & $78.6$ \\
CDANN~\citep{cdann}$^\dagger$\Tstrut & $97.1 \pm 0.3$ & $65.1 \pm 1.2$ & $70.7 \pm 0.8$ & $77.1 \pm 1.5$ & $77.5$ \\ 
MTL~\citep{mtl}$^\dagger$\Tstrut & $97.8 \pm 0.4$ & $64.3 \pm 0.3$ & $71.5 \pm 0.7$ & $75.3 \pm 1.7$ & $77.2$ \\ 
SagNet~\citep{sagnet}$^\dagger$\Tstrut & $97.9 \pm 0.4$ & $64.5 \pm 0.5$ & $71.4 \pm 1.3$ & $77.5 \pm 0.5$ & $77.8$ \\
ARM~\citep{arm}$^\dagger$\Tstrut & $98.7 \pm 0.2$ & $63.6 \pm 0.7$ & $71.3 \pm 1.2$ & $76.7 \pm 0.6$ & $77.6$ \\
VREx~\citep{vrex}$^\dagger$\Tstrut & $98.4 \pm 0.3$ & $64.4 \pm 1.4$ & $74.1 \pm 0.4$ & $76.2 \pm 1.3$ & $78.3$ \\
RSC~\citep{rsc}$^\dagger$\Tstrut & $97.9 \pm 0.1$ & $62.5 \pm 0.7$ & $72.3 \pm 1.2$ & $75.6 \pm 0.8$ & $77.1$ \\
Mixstyle~\citep{mixstyle}$^\dagger$\Tstrut & $98.6 \pm 0.3$ & $64.5 \pm 1.1$ & $72.6 \pm 0.5$ & $75.7 \pm 1.7$ & $77.9$ \\
%
FACT~\citep{fact}\Tstrut & $97.6 \pm 0.1$ & $65.5 \pm 0.5$ & $69.2 \pm 0.8$ & $73.9 \pm 0.7$ & $76.6$ \\ 
Amp-Mixup~\citep{fact}\Tstrut & $97.4 \pm 0.7$ & $65.6 \pm 0.3$ & $70.5 \pm 0.9$ & $70.1 \pm 0.8$ & $75.9$ \\ 
\hline
SWAD~\citep{swad}$^\dagger$\Tstrut & $98.8 \pm 0.1$ & $63.3 \pm 0.3$ & $75.3 \pm 0.5$ & ${\bf 79.2 \pm 0.6}$ & $79.1$ \\ 
FACT$+$SWAD\Tstrut & $98.4 \pm 0.1$ & $63.1 \pm 0.3$ & $72.4 \pm 0.5$ & $77.0 \pm 0.4$ & $77.7$ \\ 
Amp-Mixup$+$SWAD\Tstrut & $98.7 \pm 0.1$ & $63.9 \pm 0.5$ & $73.5 \pm 0.2$ & $76.7 \pm 0.2$ & $78.2$ \\ 
\hline
(Proposed) \ourModel{}\Tstrut & ${\bf 99.0 \pm 0.1}$ & $64.5 \pm 0.6$ & ${\bf 75.4 \pm 0.3}$ & $78.9 \pm 0.6$ & ${\bf 79.5}$ \\
\bottomrule
\end{tabular}
}
\end{scriptsize}
\end{table}

\begin{table}[t!]
\centering
\caption{Average accuracies on OfficeHome. The same interpretation as Table~\ref{tab:pacs}.
}
\label{tab:oh}
\begin{scriptsize}
\centering
\scalebox{0.95}{
\begin{tabular}{lccccc}
\toprule
Algorithm \ \ \ \ \ \ \ \ \ \ & \ \ \ \ \ \ \ \ C \ \ \ \ \ \ \ \ & \ \ \ \ \ \ \ \ L \ \ \ \ \ \ \ \ & \ \ \ \ \ \ \ \ S \ \ \ \ \ \ \ \ & \ \ \ \ \ \ \ \ V \ \ \ \ \ \ \ \ & \ \ Avg \ \ \\
\midrule
%
ERM~\citep{swad}$^\dagger$\Tstrut & $63.1 \pm 0.3$ & $51.9 \pm 0.4$ & $77.2 \pm 0.5$ & $78.1 \pm 0.2$ & $67.6$ \\ 
IRM~\citep{irm}$^\dagger$\Tstrut & $58.9 \pm 2.3$ & $52.2 \pm 1.6$ & $72.1 \pm 2.9$ & $74.0 \pm 2.5$ & $64.3$ \\ 
GroupDRO~\citep{group_dro}$^\dagger$\Tstrut & $60.4 \pm 0.7$ & $52.7 \pm 1.0$ & $75.0 \pm 0.7$ & $76.0 \pm 0.7$ & $66.0$ \\ 
I-Mixup~\citep{imixup_1,imixup_2,imixup_3}$^\dagger$\Tstrut & $62.4 \pm 0.8$ & $54.8 \pm 0.6$ & $76.9 \pm 0.3$ & $78.3 \pm 0.2$ & $68.1$ \\ 
MLDG~\citep{mldg}$^\dagger$\Tstrut & $61.5 \pm 0.9$ & $53.2 \pm 0.6$ & $75.0 \pm 1.2$ & $77.5 \pm 0.4$ & $66.8$ \\ 
CORAL~\citep{coral}$^\dagger$\Tstrut & $65.3 \pm 0.4$ & $54.4 \pm 0.5$ & $76.5 \pm 0.1$ & $78.4 \pm 0.5$ & $68.7$ \\ 
MMD~\citep{mmd}$^\dagger$\Tstrut & $60.4 \pm 0.2$ & $53.3 \pm 0.3$ & $74.3 \pm 0.1$ & $77.4 \pm 0.6$ & $66.4$ \\ 
DANN~\citep{dann}$^\dagger$\Tstrut & $59.9 \pm 1.3$ & $53.0 \pm 0.3$ & $73.6 \pm 0.7$ & $76.9 \pm 0.5$ & $65.9$ \\
CDANN~\citep{cdann}$^\dagger$\Tstrut & $61.5 \pm 1.4$ & $50.4 \pm 2.4$ & $74.4 \pm 0.9$ & $76.6 \pm 0.8$ & $65.7$ \\ 
MTL~\citep{mtl}$^\dagger$\Tstrut & $61.5 \pm 0.7$ & $52.4 \pm 0.6$ & $74.9 \pm 0.4$ & $76.8 \pm 0.4$ & $66.4$ \\ 
SagNet~\citep{sagnet}$^\dagger$\Tstrut & $63.4 \pm 0.2$ & $54.8 \pm 0.4$ & $75.8 \pm 0.4$ & $78.3 \pm 0.3$ & $68.1$ \\
ARM~\citep{arm}$^\dagger$\Tstrut & $58.9 \pm 0.8$ & $51.0 \pm 0.5$ & $74.1 \pm 0.1$ & $75.2 \pm 0.3$ & $64.8$ \\
VREx~\citep{vrex}$^\dagger$\Tstrut & $60.7 \pm 0.9$ & $53.0 \pm 0.9$ & $75.3 \pm 0.1$ & $76.6 \pm 0.5$ & $66.4$ \\
RSC~\citep{rsc}$^\dagger$\Tstrut & $60.7 \pm 1.4$ & $51.4 \pm 0.3$ & $74.8 \pm 1.1$ & $75.1 \pm 1.3$ & $65.5$ \\
Mixstyle~\citep{mixstyle}$^\dagger$\Tstrut & $51.1 \pm 0.3$ & $53.2 \pm 0.4$ & $68.2 \pm 0.7$ & $69.2 \pm 0.6$ & $60.4$ \\
FACT~\citep{fact}$^\dagger$\Tstrut & $60.3 \pm 0.1$ & $54.9 \pm 0.4$ & $74.5 \pm 0.1$ & $76.6 \pm 0.1$ & $66.6$ \\ 
FACT~\citep{fact}\Tstrut & $61.2 \pm 0.1$ & $55.2 \pm 0.1$ & $74.0 \pm 0.2$ & $76.2 \pm 0.4$ & $66.6$ \\ 
Amp-Mixup~\citep{fact}\Tstrut & $57.1 \pm 0.3$ & $51.9 \pm 0.1$ & $72.5 \pm 0.3$ & $74.4 \pm 0.2$ & $64.0$ \\ 
\hline
SWAD~\citep{swad}$^\dagger$\Tstrut & $66.1 \pm 0.4$ & $57.7 \pm 0.4$ & $78.4 \pm 0.1$ & $80.2 \pm 0.2$ & $70.6$ \\ 
FACT$+$SWAD\Tstrut & $66.4 \pm 0.2$ & $58.3 \pm 0.2$ & $78.0 \pm 0.1$ & $79.6 \pm 0.1$ & $70.6$ \\ 
Amp-Mixup$+$SWAD\Tstrut & $65.9 \pm 0.2$ & $57.9 \pm 0.4$ & $77.8 \pm 0.2$ & $79.7 \pm 0.1$ & $70.3$ \\ 
\hline
(Proposed) \ourModel{}\Tstrut & ${\bf 67.5 \pm 0.3}$ & ${\bf 58.5 \pm 0.1}$ & ${\bf 79.3 \pm 0.1}$ & ${\bf 80.7 \pm 0.1}$ & ${\bf 71.5}$ \\
\bottomrule
\end{tabular}
}
\end{scriptsize}
\end{table}

\begin{table}[t!]
\centering
\caption{Average accuracies on TerraIncognita. The same interpretation as Table~\ref{tab:pacs}.
}
\label{tab:ti}
\begin{scriptsize}
\centering
\scalebox{0.95}{
\begin{tabular}{lccccc}
\toprule
Algorithm \ \ \ \ \ \ \ \ \ \ \ & \ \ \ \ \ L100 \ \ \ \ \ & \ \ \ \ \ \ L38 \ \ \ \ \ \ & \ \ \ \ \ \ L43 \ \ \ \ \ \ & \ \ \ \ \ \ L46 \ \ \ \ \ \ & \ \ Avg \ \ \\
\midrule
%
ERM~\citep{swad}$^\dagger$\Tstrut & $54.3 \pm 0.4$ & $42.5 \pm 0.7$ & $55.6 \pm 0.3$ & $38.8 \pm 2.5$ & $47.8$ \\ 
IRM~\citep{irm}$^\dagger$\Tstrut & $54.6 \pm 1.3$ & $39.8 \pm 1.9$ & $56.2 \pm 1.8$ & $39.6 \pm 0.8$ & $47.6$ \\ 
GroupDRO~\citep{group_dro}$^\dagger$\Tstrut & $41.2 \pm 0.7$ & $38.6 \pm 2.1$ & $56.7 \pm 0.9$ & $36.4 \pm 2.1$ & $43.2$ \\ 
I-Mixup~\citep{imixup_1,imixup_2,imixup_3}$^\dagger$\Tstrut & $59.6 \pm 2.0$ & $42.2 \pm 1.4$ & $55.9 \pm 0.8$ & $33.9 \pm 1.4$ & $47.9$ \\ 
MLDG~\citep{mldg}$^\dagger$\Tstrut & $54.2 \pm 3.0$ & $44.3 \pm 1.1$ & $55.6 \pm 0.3$ & $36.9 \pm 2.2$ & $47.8$ \\ 
CORAL~\citep{coral}$^\dagger$\Tstrut & $51.6 \pm 2.4$ & $42.2 \pm 1.0$ & $57.0 \pm 1.0$ & $39.8 \pm 2.9$ & $47.7$ \\ 
MMD~\citep{mmd}$^\dagger$\Tstrut & $41.9 \pm 3.0$ & $34.8 \pm 1.0$ & $57.0 \pm 1.9$ & $35.2 \pm 1.8$ & $42.2$ \\ 
DANN~\citep{dann}$^\dagger$\Tstrut & $51.1 \pm 3.5$ & $40.6 \pm 0.6$ & $57.4 \pm 0.5$ & $37.7 \pm 1.8$ & $46.7$ \\
CDANN~\citep{cdann}$^\dagger$\Tstrut & $47.0 \pm 1.9$ & $41.3 \pm 4.8$ & $54.9 \pm 1.7$ & $39.8 \pm 2.3$ & $45.8$ \\ 
MTL~\citep{mtl}$^\dagger$\Tstrut & $49.3 \pm 1.2$ & $39.6 \pm 6.3$ & $55.6 \pm 1.1$ & $37.8 \pm 0.8$ & $45.6$ \\ 
SagNet~\citep{sagnet}$^\dagger$\Tstrut & $53.0 \pm 2.9$ & $43.0 \pm 2.5$ & $57.9 \pm 0.6$ & $40.4 \pm 1.3$ & $48.6$ \\
ARM~\citep{arm}$^\dagger$\Tstrut & $49.3 \pm 0.7$ & $38.3 \pm 2.4$ & $55.8 \pm 0.8$ & $38.7 \pm 1.3$ & $45.5$ \\
VREx~\citep{vrex}$^\dagger$\Tstrut & $48.2 \pm 4.3$ & $41.7 \pm 1.3$ & $56.8 \pm 0.8$ & $38.7 \pm 3.1$ & $46.4$ \\
RSC~\citep{rsc}$^\dagger$\Tstrut & $50.2 \pm 2.2$ & $39.2 \pm 1.4$ & $56.3 \pm 1.4$ & $40.8 \pm 0.6$ & $46.6$ \\
Mixstyle~\citep{mixstyle}$^\dagger$\Tstrut & $54.3 \pm 1.1$ & $34.1 \pm 1.1$ & $55.9 \pm 1.1$ & $31.7 \pm 2.1$ & $44.0$ \\
%
FACT~\citep{fact}\Tstrut & $52.4 \pm 1.2$ & $42.3 \pm 1.0$ & $55.5 \pm 0.3$ & $31.3 \pm 0.9$ & $45.4$ \\ 
Amp-Mixup~\citep{fact}\Tstrut & $56.0 \pm 0.8$ & $38.9 \pm 0.7$ & $56.9 \pm 0.2$ & $35.7 \pm 0.8$ & $46.8$ \\ 
\hline
SWAD~\citep{swad}$^\dagger$\Tstrut & $55.4 \pm 0.0$ & $44.9 \pm 1.1$ & $59.7 \pm 0.4$ & $39.9 \pm 0.2$ & $50.0$ \\ 
FACT$+$SWAD\Tstrut & $57.0 \pm 0.6$ & ${\bf 46.6 \pm 1.1}$ & ${\bf 60.3 \pm 0.5}$ & $40.1 \pm 0.3$ & $51.0$ \\ 
Amp-Mixup$+$SWAD\Tstrut & $56.6 \pm 0.6$ & $46.3 \pm 0.3$ & $60.2 \pm 0.6$ & $41.8 \pm 0.4$ & $51.2$ \\ 
\hline
(Proposed) \ourModel{}\Tstrut & ${\bf 61.0 \pm 0.3}$ & $46.2 \pm 2.3$ & ${\bf 60.3 \pm 0.7}$ & ${\bf 42.3 \pm 0.9}$ & ${\bf 52.4}$ \\
\bottomrule
\end{tabular}
}
\end{scriptsize}
\end{table}

\begin{table}[t!]
\centering
\caption{Average accuracies on DomainNet. The same interpretation as Table~\ref{tab:pacs}.
}
\label{tab:dn}
\begin{scriptsize}
\centering
\scalebox{0.95}{
\begin{tabular}{lccccccc}
\toprule
Algorithm \ \ \ \ \ \ \ \ \ & \ \ \ \ \ \ \ C \ \ \ \ \ \ & \ \ \ \ \ \ \ I \ \ \ \ \ \ & \ \ \ \ \ \ \ P \ \ \ \ \ \ & \ \ \ \ \ \ \ Q \ \ \ \ \ \ & \ \ \ \ \ \ \ R \ \ \ \ \ \ & \ \ \ \ \ \ \ S \ \ \ \ \ \ & \ \ Avg \ \ \\
\midrule
%
DMG~\citep{dmg}$^\dagger$\Tstrut & $65.2$ & $22.2$ & $50.0$ & $15.7$ & $59.6$ & $49.0$ & $43.6$ \\ 
MetaReg~\citep{metareg}$^\dagger$\Tstrut & $59.8$ & ${\bf 25.6}$ & $50.2$ & $11.5$ & $64.6$ & $50.1$ & $43.6$ \\ 
%
ERM~\citep{swad}$^\dagger$\Tstrut & $63.0 \pm 0.2$ & $21.2 \pm 0.2$ & $50.1 \pm 0.4$ & $13.9 \pm 0.5$ & $63.7 \pm 0.2$ & $52.0 \pm 0.5$ & $44.0$ \\ 
IRM~\citep{irm}$^\dagger$\Tstrut & $48.5 \pm 2.8$ & $15.0 \pm 1.5$ & $38.3 \pm 4.3$ & $10.9 \pm 0.5$ & $48.2 \pm 5.2$ & $42.3 \pm 3.1$ & $33.9$ \\ 
GroupDRO~\citep{group_dro}$^\dagger$\Tstrut & $47.2 \pm 0.5$ & $17.5 \pm 0.4$ & $33.8 \pm 0.5$ & $9.3 \pm 0.3$ & $51.6 \pm 0.4$ & $40.1 \pm 0.6$ & $33.3$ \\ 
I-Mixup~(Citation as before)\Tstrut
& $55.7 \pm 0.3$ & $18.5 \pm 0.5$ & $44.3 \pm 0.5$ & $12.5 \pm 0.4$ & $55.8 \pm 0.3$ & $48.2 \pm 0.5$ & $39.2$ \\ 
MLDG~\citep{mldg}$^\dagger$\Tstrut & $59.1 \pm 0.2$ & $19.1 \pm 0.3$ & $45.8 \pm 0.7$ & $13.4 \pm 0.3$ & $59.6 \pm 0.2$ & $50.2 \pm 0.4$ & $41.2$ \\ 
CORAL~\citep{coral}$^\dagger$\Tstrut & $59.2 \pm 0.1$ & $19.7 \pm 0.2$ & $46.6 \pm 0.3$ & $13.4 \pm 0.4$ & $59.8 \pm 0.2$ & $50.1 \pm 0.6$ & $41.5$ \\ 
MMD~\citep{mmd}$^\dagger$\Tstrut & $32.1 \pm 13.3$ & $11.0 \pm 4.6$ & $26.8 \pm 11.3$ & $8.7 \pm 2.1$ & $32.7 \pm 13.8$ & $28.9 \pm 11.9$ & $23.4$ \\ 
DANN~\citep{dann}$^\dagger$\Tstrut & $53.1 \pm 0.2$ & $18.3 \pm 0.1$ & $44.2 \pm 0.7$ & $11.8 \pm 0.1$ & $55.5 \pm 0.4$ & $46.8 \pm 0.6$ & $38.3$ \\
CDANN~\citep{cdann}$^\dagger$\Tstrut & $54.6 \pm 0.4$ & $17.3 \pm 0.1$ & $43.7 \pm 0.9$ & $12.1 \pm 0.7$ & $56.2 \pm 0.4$ & $45.9 \pm 0.5$ & $38.3$ \\ 
MTL~\citep{mtl}$^\dagger$\Tstrut & $57.9 \pm 0.5$ & $18.5 \pm 0.4$ & $46.0 \pm 0.1$ & $12.5 \pm 0.1$ & $59.5 \pm 0.3$ & $49.2 \pm 0.1$ & $40.6$ \\ 
SagNet~\citep{sagnet}$^\dagger$\Tstrut & $57.7 \pm 0.3$ & $19.0 \pm 0.2$ & $45.3 \pm 0.3$ & $12.7 \pm 0.5$ & $58.1 \pm 0.5$ & $48.8 \pm 0.2$ & $40.3$ \\
ARM~\citep{arm}$^\dagger$\Tstrut & $49.7 \pm 0.3$ & $16.3 \pm 0.5$ & $40.9 \pm 1.1$ & $9.4 \pm 0.1$ & $53.4 \pm 0.4$ & $43.5 \pm 0.4$ & $35.5$ \\
VREx~\citep{vrex}$^\dagger$\Tstrut & $47.3 \pm 3.5$ & $16.0 \pm 1.5$ & $35.8 \pm 4.6$ & $10.9 \pm 0.3$ & $49.6 \pm 4.9$ & $42.0 \pm 3.0$ & $33.6$ \\
RSC~\citep{rsc}$^\dagger$\Tstrut & $55.0 \pm 1.2$ & $18.3 \pm 0.5$ & $44.4 \pm 0.6$ & $12.2 \pm 0.2$ & $55.7 \pm 0.7$ & $47.8 \pm 0.9$ & $38.9$ \\
Mixstyle~\citep{mixstyle}$^\dagger$\Tstrut & $51.9 \pm 0.4$ & $13.3 \pm 0.2$ & $37.0 \pm 0.5$ & $12.3 \pm 0.1$ & $46.1 \pm 0.3$ & $43.4 \pm 0.4$ & $34.0$ \\
%
FACT~\citep{fact}\Tstrut & $62.5 \pm 0.3$ & $19.4 \pm 0.1$ & $48.2 \pm 0.4$ & $13.9 \pm 0.3$ & $60.5 \pm 0.7$ & $51.0 \pm 0.7$ & $42.6$ \\ 
Amp-Mixup~\citep{fact}\Tstrut & $62.3 \pm 0.1$ & $19.0 \pm 0.2$ & $47.2 \pm 0.4$ & $12.9 \pm 0.6$ & $59.5 \pm 0.3$ & $51.0 \pm 0.1$ & $42.0$ \\ 
\hline
SWAD~\citep{swad}$^\dagger$\Tstrut & $66.0 \pm 0.1$ & $22.4 \pm 0.3$ & $53.5 \pm 0.1$ & $16.1 \pm 0.2$ & $65.8 \pm 0.4$ & $55.5 \pm 0.3$ & $46.5$ \\ 
FACT$+$SWAD\Tstrut & $66.3 \pm 0.1$ & $22.7 \pm 0.2$ & $53.7 \pm 0.1$ & $16.3 \pm 0.1$ & $65.0 \pm 0.6$ & $55.9 \pm 0.1$ & $46.7$ \\ 
Amp-Mixup$+$SWAD\Tstrut & $66.1 \pm 0.1$ & $22.4 \pm 0.2$ & $53.3 \pm 0.1$ & $16.2 \pm 0.3$ & $64.6 \pm 0.5$ & $55.6 \pm 0.1$ & $46.4$ \\ 
\hline
(Proposed) \ourModel{}\Tstrut & ${\bf 66.7 \pm 0.1}$ & $22.9 \pm 0.2$ & ${\bf 54.0 \pm 0.1}$ & ${\bf 16.7 \pm 0.2}$ & ${\bf 65.9 \pm 0.1}$ & ${\bf 56.3 \pm 0.1}$ & ${\bf 47.1}$ \\
\bottomrule
\end{tabular}
}
\end{scriptsize}
\end{table}


\begin{table}
\caption{Sensitivity analysis on the SMCD loss trade off $\eta$ on PACS and OfficeHome. 
}
\vspace{+0.3em}
\centering
\begin{scriptsize}
\centering
%
(a) PACS \\
\begin{tabular}{lccccc}
\toprule
 & Art & Cartoon & Photo & Sketch & Average
\\ \hline
$\eta=0.0$\Tstrut \ \ & $89.08 \pm 0.14$ & $83.55 \pm 0.16$ & $97.23 \pm 0.19$ & $82.55 \pm 0.30$ & $88.10$ \\ 
$\eta=0.01$\Tstrut \ \ & $89.24 \pm 0.26$ & $84.41 \pm 0.53$ & $97.17 \pm 0.07$ & $84.31 \pm 0.37$ & $88.78$ \\ 
$\eta=0.05$\Tstrut \ \ & $89.52 \pm 0.26$ & $84.83 \pm 0.20$ & $97.33 \pm 0.17$ & $83.75 \pm 0.14$ & $88.86$ \\ 
$\eta=0.1$\Tstrut \ \ & $\ {\bf 89.80 \pm 0.34} \ $ & $\ {\bf 85.16 \pm 0.65} \ $ & $\ {\bf 97.59 \pm 0.27} \ $ & $\ {\bf 84.67 \pm 0.82} \ $ & $\ {\bf 89.30} \ $ \\ 
$\eta=0.2$\Tstrut \ \ & $89.40 \pm 0.60$ & $84.57 \pm 0.25$ & $97.33 \pm 0.15$ & $83.88 \pm 0.16$ & $88.80$ \\ 
$\eta=0.5$\Tstrut \ \ & $89.00 \pm 0.14$ & $84.40 \pm 0.38$ & $97.03 \pm 0.17$ & $83.10 \pm 0.80$ & $88.38$ \\ 
$\eta=1.0$\Tstrut \ \ & $89.11 \pm 0.36$ & $84.20 \pm 0.55$ & $96.49 \pm 0.22$ & $82.39 \pm 0.57$ & $88.05$ \\
\bottomrule
\end{tabular}
%
\\
\vspace{+0.5em}
(b) OfficeHome \\
\begin{tabular}{lccccc}
\toprule
 & Art & Clipart & Product & Real & Average
\\ \hline
$\eta=0.0$\Tstrut \ \ & $66.09 \pm 0.28$ & $57.72 \pm 0.34$ & $78.47 \pm 0.16$ & $80.19 \pm 0.11$ & $70.62$ \\ 
$\eta=0.01$\Tstrut \ \ & $66.86 \pm 0.17$ & $58.43 \pm 0.34$ & $78.53 \pm 0.09$ & $80.51 \pm 0.31$ & $71.08$ \\ 
$\eta=0.05$\Tstrut \ \ & $66.95 \pm 0.09$ & ${\bf 58.56 \pm 0.32}$ & $78.96 \pm 0.28$ & $80.46 \pm 0.24$ & $71.23$ \\ 
$\eta=0.1$\Tstrut \ \ & $\ {\bf 67.46 \pm 0.28} \ $ & $58.45 \pm 0.13$ & $\ {\bf 79.27 \pm 0.07} \ $ & $\ {\bf 80.70 \pm 0.11} \ $ & $\ {\bf 71.47} \ $ \\ 
$\eta=0.15$\Tstrut \ \ & $66.46 \pm 0.33$ & $58.31 \pm 0.22$ & $78.59 \pm 0.33$ & $80.51 \pm 0.09$ & $70.97$ \\ 
$\eta=0.2$\Tstrut \ \ & $66.05 \pm 0.03$ & $58.10 \pm 0.11$ & $78.69 \pm 0.18$ & $80.45 \pm 0.06$ & $70.82$ \\ 
$\eta=0.25$\Tstrut \ \ & $66.15 \pm 0.17$ & $58.29 \pm 0.36$ & $78.16 \pm 0.04$ & $79.87 \pm 0.21$ & $70.62$ \\ 
%
%
\bottomrule
\end{tabular}
%
\end{scriptsize}
\label{tab:abl_eta}
\end{table}

\begin{table}
\caption{Sensitivity analysis on the post-mixup trade off $\alpha$ on PACS and OfficeHome. 
}
\vspace{+0.3em}
\centering
\begin{scriptsize}
\centering
%
(a) PACS \\
\begin{tabular}{lccccc}
\toprule
 & Art & Cartoon & Photo & Sketch & Average
\\ \hline
$\alpha=0.0$\Tstrut \ \ & $89.29 \pm 0.37$ & $83.55 \pm 0.20$ & $97.11 \pm 0.07$ & $82.17 \pm 0.91$ & $88.03$ \\ 
$\alpha=0.2$\Tstrut \ \ & $89.23 \pm 0.22$ & $83.80 \pm 0.21$ & $97.25 \pm 0.05$ & $82.59 \pm 0.84$ & $88.22$ \\ 
$\alpha=0.4$\Tstrut \ \ & $89.37 \pm 0.17$ & $83.93 \pm 0.07$ & $97.27 \pm 0.10$ & $83.37 \pm 0.43$ & $88.49$ \\ 
$\alpha=0.6$\Tstrut \ \ & $89.42 \pm 0.48$ & $84.30 \pm 0.16$ & $97.31 \pm 0.13$ & $83.71 \pm 0.60$ & $88.69$ \\ 
$\alpha=0.8$\Tstrut \ \ & $89.47 \pm 0.61$ & $84.39 \pm 0.18$ & $97.41 \pm 0.15$ & $83.68 \pm 0.15$ & $88.74$ \\ 
$\alpha=0.9$\Tstrut \ \ & $89.66 \pm 0.23$ & $85.04 \pm 0.28$ & $\ {\bf 97.60 \pm 0.13} \ $ & $84.27 \pm 0.40$ & $89.14$ \\ 
$\alpha=1.0$\Tstrut \ \ & $\ {\bf 89.80 \pm 0.34} \ $ & $\ {\bf 85.16 \pm 0.65} \ $ & $97.59 \pm 0.27$ & $\ {\bf 84.67 \pm 0.82} \ $ & $\ {\bf 89.30} \ $ \\
\bottomrule
\end{tabular}
%
\\
\vspace{+0.5em}
(b) OfficeHome \\
\begin{tabular}{lccccc}
\toprule
 & Art & Clipart & Product & Real & Average
\\ \hline
$\alpha=0.0$\Tstrut \ \ & $65.99 \pm 0.17$ & $57.72 \pm 0.16$ & $78.36 \pm 0.08$ & $80.22 \pm 0.09$ & $70.57$ \\ 
$\alpha=0.1$\Tstrut \ \ & $67.04 \pm 0.15$ & $58.09 \pm 0.11$ & $78.72 \pm 0.15$ & $80.36 \pm 0.08$ & $71.05$ \\ 
$\alpha=0.2$\Tstrut \ \ & $\ {\bf 67.46 \pm 0.28} \ $ & $58.45 \pm 0.13$ & $\ {\bf 79.27 \pm 0.07} \ $ & $\ {\bf 80.70 \pm 0.11} \ $ & $\ {\bf 71.47} \ $ \\ 
$\alpha=0.3$\Tstrut \ \ & $66.98 \pm 0.24$ & $\ {\bf 58.50 \pm 0.22} \ $ & $78.87 \pm 0.18$ & $80.45 \pm 0.06$ & $71.20$ \\ 
$\alpha=0.4$\Tstrut \ \ & $67.04 \pm 0.32$ & $58.37 \pm 0.26$ & $78.57 \pm 0.09$ & $80.43 \pm 0.09$ & $71.10$ \\ 
$\alpha=0.6$\Tstrut \ \ & $66.45 \pm 0.33$ & $57.99 \pm 0.20$ & $78.53 \pm 0.21$ & $80.22 \pm 0.11$ & $70.80$ \\ 
$\alpha=0.8$\Tstrut \ \ & $66.56 \pm 0.23$ & $57.89 \pm 0.18$ & $78.50 \pm 0.12$ & $80.21 \pm 0.18$ & $70.79$ \\ 
$\alpha=1.0$\Tstrut \ \ & $66.47 \pm 0.30$ & $57.96 \pm 0.09$ & $78.46 \pm 0.30$ & $79.97 \pm 0.29$ & $70.72$ \\
\bottomrule
\end{tabular}
%
\end{scriptsize}
\label{tab:abl_alpha}
\end{table}

\subsection{Derivation of ELBO in Variational Inference}\label{sec:derivations}

We derive the evidence lower bound (ELBO) in (\ref{eq:elbo}) in the main paper. To enforce $Q_\lambda(W) \approx P(W|S,\theta)$, we minimise their KL divergence,
\begin{align}
&\textrm{KL} \big( Q_\lambda(W) || P(W|S,\theta) \big) \ = \ 
\mathbb{E}_{Q_\lambda(W)} \bigg[ \log \frac{Q_\lambda(W)}{P(W|S,\theta)} \bigg] \label{eq:elbo_deriv_1} \\
& \ \ = \ \mathbb{E}_{Q_\lambda(W)} \bigg[ \log \frac{Q_\lambda(W) P(S|\theta)}{P(S|W,\theta) P(W)} \bigg] \label{eq:elbo_deriv_2} \\
& \ \ = \ \log P(S|\theta) - \mathbb{E}_{Q_\lambda(W)} \big[ \log P(S|W,\theta) \big] + \mathbb{E}_{Q_\lambda(W)} \bigg[ \log \frac{Q_\lambda(W)}{P(W)} \bigg] \label{eq:elbo_deriv_3} \\
& \ \ = \ \log P(S|\theta) - \mathbb{E}_{Q_\lambda(W)} \big[ \log P(S|W,\theta) \big] + \textrm{KL} \big( Q_\lambda(W) || P(W) \big). \label{eq:elbo_deriv_4} \\
& \ \ = \ \log P(S|\theta) - \sum_{(x,y)\sim S} \mathbb{E}_{Q_\lambda(W)} \big[ \log P(y|x,W,\theta) \big] + \textrm{KL} \big( Q_\lambda(W) || P(W) \big). \label{app_eq:elbo_deriv_5}
\end{align}
Since KL divergence is non-negative, re-arranging (\ref{app_eq:elbo_deriv_5}) yields:
\begin{equation}
\log P(S|\theta) \geq 
  \sum_{(x,y)\sim S} \mathbb{E}_{Q_\lambda(W)} \big[ 
    \log P(y|x,W,\theta) \big]  - 
  \textrm{KL} \big( Q_\lambda(W) || P(W) \big),
\label{app_eq:elbo}
\end{equation}
and the right hand side constitutes the ELBO.

\subsection{Additional Experimental Results}\label{sec:extra_expmt}

\subsubsection{Results on ResNet-18 Backbone}\label{sec:resnet18_pacs}

To test our approach on backbone networks other than ResNet-50, we run experiments with the ResNet-18 backbone on the PACS dataset. The results are summarised in Table~\ref{tab:resnet18_pacs}. Compared to the recent approaches MixStyle~\citep{mixstyle} and EFDMix~\citep{efdmix}, our approach AGFA again shows higher performance even with the smaller ResNet-18 backbone.

\begin{table}[t!]
\centering
\caption{Average accuracies on PACS with ResNet-18 backbone. Results on ERM, Mixup~\citep{mixup},  MixStyle~\citep{mixstyle}, and EFDMix~\citep{efdmix} are excerpted from~\citep{efdmix}.
}
\vspace{+0.2em}
\label{tab:resnet18_pacs}
\begin{footnotesize}
\centering
\scalebox{0.95}{
\begin{tabular}{lccccc}
\toprule
Algorithm \ \ \ \ \ \ \ \ \ \ \ \ \ \ & \ \ \ \ \ \ Art \ \ \ \ \ \ & \ \ \ \ \ \ Cartoon \ \ \ \ \ \ & \ \ \ \ \ \ Painting \ \ \ \ \ \ & \ \ \ \ \ \ Sketch \ \ \ \ \ \ & \ \ Avg \ \ \\
\midrule
ERM\Tstrut & $77.0 \pm 0.6$ & $75.9 \pm 0.6$ & $96.0 \pm 0.1$ & $69.2 \pm 0.6$ & $79.5$ \\ 
Mixup\Tstrut & $76.8 \pm 0.7$ & $74.9 \pm 0.7$ & $95.8 \pm 0.3$ & $66.6 \pm 0.7$ & $78.5$ \\ 
MixStyle\Tstrut & $83.1 \pm 0.8$ & $78.6 \pm 0.9$ & $95.9 \pm 0.4$ & $74.2 \pm 2.7$ & $82.9$ \\ 
EFDMix\Tstrut & $83.9 \pm 0.4$ & ${\bf 79.4 \pm 0.7}$ & ${\bf 96.8 \pm 0.4}$ & $75.0 \pm 0.7$ & $83.9$ \\ 
\hline
(Proposed) \ourModel{}\Tstrut & ${\bf 84.5 \pm 0.6}$ & $78.5 \pm 0.5$ & $95.7 \pm 0.1$ & ${\bf 80.9 \pm 0.2}$ & ${\bf 84.9}$ \\
\bottomrule
\end{tabular}
}
\end{footnotesize}
\end{table}

\subsubsection{Results on Coloured-MNIST and Rotated-MNIST}\label{sec:crmnist}

Although relatively smaller and easier datasets in the DomainBed benchmark, we also test our method on the Coloured-MNIST and Rotated-MNIST datasets. Following the experimental protocols including the four-layer ConvNet backbone as in~\citep{domainbed}, the test accuracies are reported in Table~\ref{tab:cmnist} (Colored-MNIST) and Table~\ref{tab:rmnist} (Rotated-MNIST). As shown, all approaches including ours perform equally well on these datasets.

\begin{table}[t!]
\centering
\caption{Average accuracies on Colored-MNIST with the four-layer ConvNet backbone. Results on competing methods are excerpted from~\citep{domainbed}.
}
\vspace{+0.2em}
\label{tab:cmnist}
\begin{footnotesize}
\centering
\scalebox{0.95}{
\begin{tabular}{lcccc}
\toprule
Algorithm & 0.1 & 0.2 & 0.9 & Avg \\
\midrule
ERM\Tstrut & $72.7 \pm 0.2$ & $73.2 \pm 0.3$ & $10.0 \pm 0.0$ & $52.0$ \\ 
IRM\Tstrut & $72.0 \pm 0.2$ & $73.2 \pm 0.0$ & $10.1 \pm 0.2$ & $51.8$ \\ 
DRO\Tstrut & $72.7 \pm 0.3$ & $73.1 \pm 0.3$ & $10.0 \pm 0.0$ & $51.9$ \\ 
Mixup\Tstrut & $72.4 \pm 0.2$ & $73.3 \pm 0.3$ & $10.0 \pm 0.1$ & $51.9$ \\ 
MLDG\Tstrut & $71.4 \pm 0.4$ & $73.3 \pm 0.0$ & $10.0 \pm 0.1$ & $51.6$ \\ 
CORAL\Tstrut & $71.8 \pm 0.4$ & $73.3 \pm 0.2$ & $10.1 \pm 0.1$ & $51.7$ \\ 
MMD\Tstrut & $72.1 \pm 0.2$ & $72.8 \pm 0.2$ & $10.5 \pm 0.2$ & $51.8$ \\ 
ADA\Tstrut & $72.0 \pm 0.3$ & $72.4 \pm 0.5$ & $10.0 \pm 0.2$ & $51.5$ \\ 
CondADA\Tstrut & $72.2 \pm 0.3$ & $73.2 \pm 0.2$ & $10.4 \pm 0.3$ & $51.9$ \\ 
\hline
(Proposed) \ourModel{}\Tstrut & $72.6 \pm 0.1$ & $73.8 \pm 0.1$ & $10.5 \pm 0.1$ & $52.3$ \\
\bottomrule
\end{tabular}
}
\end{footnotesize}
\end{table}

\begin{table}[t!]
\centering
\caption{Average accuracies on Rotated-MNIST with the four-layer ConvNet backbone. Results on competing methods are excerpted from~\citep{domainbed}.
}
\vspace{+0.2em}
\label{tab:rmnist}
\begin{footnotesize}
\centering
\scalebox{0.95}{
\begin{tabular}{lccccccc}
\toprule
Algorithm & 0 & 15 & 30 & 45 & 60 & 75 & Avg \\
\midrule
ERM\Tstrut & $95.6 \pm 0.1$ & $99.0 \pm 0.1$ & $98.9 \pm 0.0$ & $99.1 \pm 0.1$ & $99.0 \pm 0.0$ & $96.7 \pm 0.2$ & $98.1$ \\ 
IRM\Tstrut & $95.9 \pm 0.2$ & $98.9 \pm 0.0$ & $99.0 \pm 0.0$ & $98.8 \pm 0.1$ & $98.9 \pm 0.1$ & $95.5 \pm 0.3$ & $97.8$ \\ 
DRO\Tstrut & $95.9 \pm 0.1$ & $98.9 \pm 0.0$ & $99.0 \pm 0.1$ & $99.0 \pm 0.0$ & $99.0 \pm 0.0$ & $96.9 \pm 0.1$ & $98.1$ \\ 
Mixup\Tstrut & $96.1 \pm 0.2$ & $99.1 \pm 0.0$ & $98.9 \pm 0.0$ & $99.0 \pm 0.0$ & $99.0 \pm 0.1$ & $96.6 \pm 0.1$ & $98.1$ \\ 
MLDG\Tstrut & $95.9 \pm 0.2$ & $98.9 \pm 0.1$ & $99.0 \pm 0.0$ & $99.1 \pm 0.0$ & $99.0 \pm 0.0$ & $96.0 \pm 0.2$ & $98.0$ \\ 
CORAL\Tstrut & $95.7 \pm 0.2$ & $99.0 \pm 0.0$ & $99.1 \pm 0.1$ & $99.1 \pm 0.0$ & $99.0 \pm 0.0$ & $96.7 \pm 0.2$ & $98.1$ \\ 
MMD\Tstrut & $96.6 \pm 0.1$ & $98.9 \pm 0.0$ & $98.9 \pm 0.1$ & $99.1 \pm 0.1$ & $99.0 \pm 0.0$ & $96.2 \pm 0.1$ & $98.1$ \\ 
DANN\Tstrut & $95.6 \pm 0.3$ & $98.9 \pm 0.0$ & $98.9 \pm 0.0$ & $99.0 \pm 0.1$ & $98.9 \pm 0.0$ & $95.9 \pm 0.5$ & $97.9$ \\ 
C-DANN\Tstrut & $96.0 \pm 0.5$ & $98.8 \pm 0.0$ & $99.0 \pm 0.1$ & $99.1 \pm 0.0$ & $98.9 \pm 0.1$ & $96.5 \pm 0.3$ & $98.0$ \\ 
\hline
(Proposed) \ourModel{}\Tstrut & $98.1 \pm 0.1$ & $98.9 \pm 0.0$ & $99.0 \pm 0.0$ & $98.8 \pm 0.0$ & $99.0 \pm 0.0$ & $96.4 \pm 0.1$ & $98.0$ \\
\bottomrule
\end{tabular}
}
\end{footnotesize}
\end{table}

\subsubsection{Results on Single-Source Generalisation}\label{sec:single_source}

We have focused predominantly on the most popular leave-one-domain-out DG setting in our empirical study. Another reasonable experimental setting is single source generalisation setting: training on only one source domain and testing on the rest domains. Our single source domain results on the PACS benchmark are shown in Table~\ref{tab:ss_pacs} for (a) ResNet-18 and (b) ResNet-50 backbones. The results indicate that improvement of the proposed AGFA over the existing DG methods is even more pronounced: averaged accuracies higher than the best prior method EFDMIX~\citep{efdmix} by about $10\%$ for ResNet-18 and by about $7\%$ for ResNet-50.

\begin{table}[t!]
\centering
\caption{Single source domain generalisation results on PACS with (a) ResNet-18 and (b) ResNet-50 backbones. Each column shows  test accuracies averaged over the rest three target domains. Results on ERM, MixStyle~\citep{mixstyle}, and EFDMix~\citep{efdmix} are excerpted from~\citep{efdmix}.
}
\vspace{+0.2em}
\label{tab:ss_pacs}
\begin{footnotesize}
\centering
(a) ResNet-18 \\
\scalebox{0.95}{
\begin{tabular}{lccccc}
\toprule
Algorithm \ \ \ \ \ \ \ \ \ \ \ \ \ \ & \ \ \ \ \ \ Art \ \ \ \ \ \ & \ \ \ \ \ \ Cartoon \ \ \ \ \ \ & \ \ \ \ \ \ Painting \ \ \ \ \ \ & \ \ \ \ \ \ Sketch \ \ \ \ \ \ & \ \ Avg \ \ \\
\midrule
ERM\Tstrut & $58.6 \pm 2.4$ & $66.4 \pm 0.7$ & $34.0 \pm 1.8$ & $27.5 \pm 4.3$ & $46.6$ \\ 
MixStyle\Tstrut & $61.9 \pm 2.2$ & $71.5 \pm 0.8$ & $41.2 \pm 1.8$ & $32.2 \pm 4.1$ & $51.7$ \\ 
EFDMix\Tstrut & $63.2 \pm 2.3$ & $73.9 \pm 0.7$ & $42.5 \pm 1.8$ & $38.1 \pm 3.7$ & $54.4$ \\ 
\hline
(Proposed) \ourModel{}\Tstrut & ${\bf 74.2 \pm 1.1}$ & ${\bf 77.5 \pm 0.6}$ & ${\bf 48.5 \pm 2.6}$ & ${\bf 58.3 \pm 0.9}$ & ${\bf 64.6}$ \\
\bottomrule
\end{tabular}
}
\\
\vspace{+0.5em}
(b) ResNet-50 \\
\scalebox{0.95}{
\begin{tabular}{lccccc}
\toprule
Algorithm \ \ \ \ \ \ \ \ \ \ \ \ \ \ & \ \ \ \ \ \ Art \ \ \ \ \ \ & \ \ \ \ \ \ Cartoon \ \ \ \ \ \ & \ \ \ \ \ \ Painting \ \ \ \ \ \ & \ \ \ \ \ \ Sketch \ \ \ \ \ \ & \ \ Avg \ \ \\
\midrule
ERM\Tstrut & $63.5 \pm 1.3$ & $69.2 \pm 1.6$ & $38.0 \pm 0.9$ & $31.4 \pm 1.5$ & $50.5$ \\ 
MixStyle\Tstrut & $73.2 \pm 1.1$ & $74.8 \pm 1.1$ & $46.0 \pm 2.0$ & $40.6 \pm 2.0$ & $58.6$ \\ 
EFDMix\Tstrut & $75.3 \pm 0.9$ & $77.4 \pm 0.8$ & $48.0 \pm 0.9$ & $44.2 \pm 2.4$ & $61.2$ \\ 
\hline
(Proposed) \ourModel{}\Tstrut & ${\bf 79.8 \pm 0.9}$ & ${\bf 81.7 \pm 0.6}$ & ${\bf 48.6 \pm 0.5}$ & ${\bf 64.6 \pm 1.1}$ & ${\bf 68.7}$ \\
\bottomrule
\end{tabular}
}
\end{footnotesize}
\end{table}

\subsubsection{Comparison with Pixel-based Target Image Generation}\label{sec:pixel_based}

\begin{figure}[t!]
\begin{center}
%
\centering
\includegraphics[trim = 2mm 2mm 6mm 4mm, clip, scale=0.505 
]{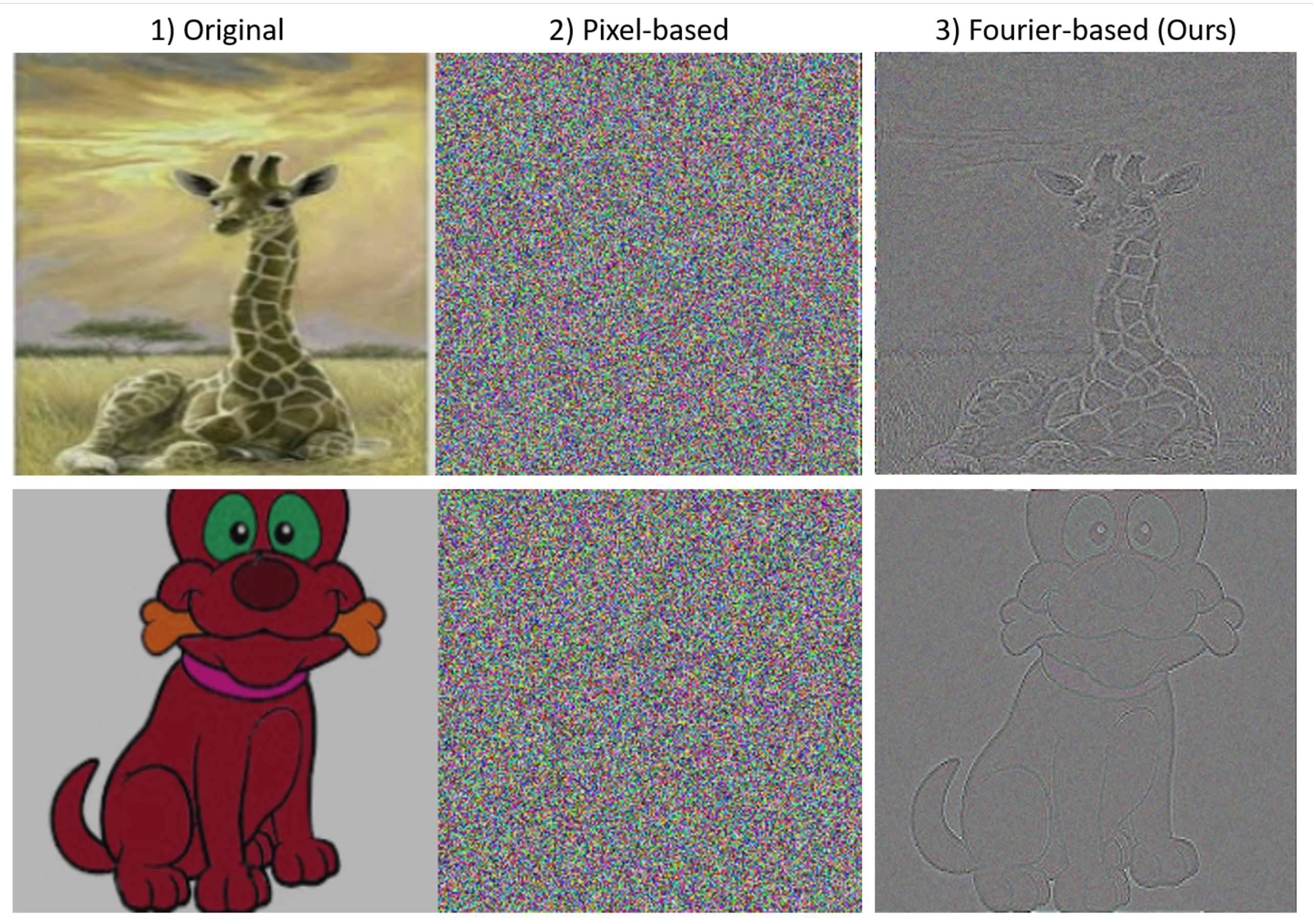}
%
\end{center}
\vspace{-1.0em}
\caption{
Comparison between pixel-based and our Fourier-based generated target images. Whereas the pixel-based generation is visually uninformative and looks like pure random noise, our Fourier-based generation contains salient object edge information that is closely related to class semantics.
}
\label{fig:vis_acps_pixel_based}
\end{figure}

Our Fourier-based target image generation is effective for preserving semantic class information from the source domains, thanks to the phase/amplitude separation. To see if non-Fourier-based generation also has similar property, we visualise adversarial target images generated by a purely pixel-based manner without phase/amplitude separation. For the linear pixel-based generator model (from $100$-dim input noise to full image pixels), which performed slightly better than nonlinear ones in test accuracy, we show some examples in Fig.~\ref{fig:vis_acps_pixel_based}. 
Whereas the pixel-based generation is visually uninformative and looks like pure random noise, our Fourier-based generation contains salient object edge information that is closely related to class semantics.

\end{document}